\documentclass[11pt]{article}

\usepackage[preprint]{acl}

\usepackage{times}
\usepackage{latexsym}

\usepackage[T1]{fontenc}

\usepackage[utf8]{inputenc}

\usepackage{microtype}

\usepackage{inconsolata}

\usepackage{graphicx}

\usepackage{newfloat}
\usepackage{listings}
\usepackage{tabularx}
\usepackage{multirow}
\usepackage{booktabs}
\usepackage{tabularx}
\usepackage{makecell}
\usepackage{caption}
\usepackage{colortbl}
\usepackage{float}
\usepackage[table]{xcolor}
\usepackage{xcolor}
\usepackage{hyperref}
\usepackage{url}
\usepackage{arydshln}
\usepackage{array}

\title{MedErrBench: A Fine-Grained Multilingual Benchmark for Medical Error Detection and Correction with Clinical Expert Annotations}

\author{
 \textbf{Congbo Ma\textsuperscript{1}},
 \textbf{Yichun Zhang\textsuperscript{2}},
 \textbf{Yousef Al-Jazzazi\textsuperscript{1}},
 \textbf{Ahamed Foisal\textsuperscript{1}},
 \textbf{Laasya Sharma\textsuperscript{3}},
 \\
 \textbf{Yousra Sadqi\textsuperscript{4}},
 \textbf{Khaled Saleh\textsuperscript{4}},
 \textbf{Jihad Mallat\textsuperscript{4}},
 \textbf{Farah E. Shamout\textsuperscript{1}}
\\
 \textsuperscript{1}New York University Abu Dhabi,
 \textsuperscript{2}New York University,
 \\
 \textsuperscript{3}University of Birmingham,
 \textsuperscript{4}Cleveland Clinic Abu Dhabi
}

\begin{document}
\maketitle
\begin{abstract}

Inaccuracies in existing or generated clinical text may lead to serious adverse consequences, especially if it is a misdiagnosis or incorrect treatment suggestion. With Large Language Models (LLMs) increasingly being used across diverse healthcare applications, comprehensive evaluation through dedicated benchmarks is crucial. However, such datasets remain scarce, especially across diverse languages and contexts. In this paper, we introduce MedErrBench, the first multilingual benchmark for error detection, localization, and correction, developed under the guidance of experienced clinicians. Based on an expanded taxonomy of ten common error types, MedErrBench covers English, Arabic and Chinese, with natural clinical cases annotated and reviewed by domain experts.
We assessed the performance of a range of general-purpose, language-specific, and medical-domain language models across all three tasks. Our results reveal notable performance gaps, particularly in non-English settings, highlighting the need for clinically grounded, language-aware systems. By making MedErrBench and our evaluation protocols publicly-available, we aim to advance multilingual clinical NLP to promote safer and more equitable AI-based healthcare globally. The dataset is available in the supplementary material. An anonymized version of the dataset is available at: \url{https://github.com/congboma/MedErrBench}.
\end{abstract}

\section{Introduction}

\begin{figure}[t]
     \centering
\includegraphics[width=0.48\textwidth]{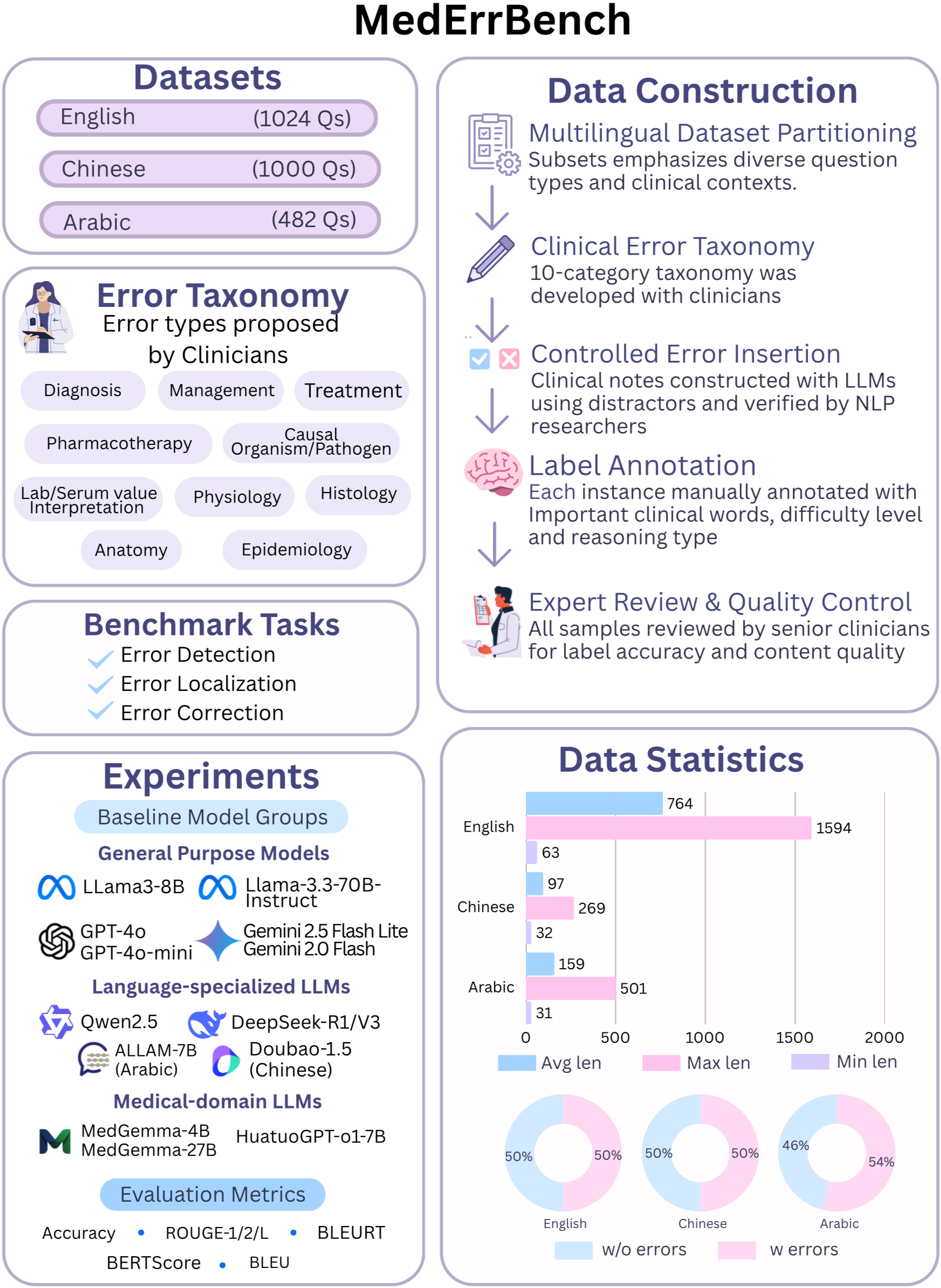}
    \caption{Overview of MedErrBench.}
    \label{fig: overview}
    \vspace{-6mm}
\end{figure}

Medical error detection and correction are essential for ensuring patient safety and healthcare quality \cite{anjum2024patient, ahsani2022interventions}. Some errors, such as misdiagnoses, can lead to severe adverse outcomes, such as morbidity and mortality, and high economic costs \cite{newman2021rate, soori2024errors}. This need is especially critical in the era of generative AI and Large Language Models (LLMs).
Despite their importance, automated methods for detecting and correcting medical errors remain underexplored due to three main challenges. First, there is a significant shortage of publicly available datasets for medical error correction and detection. To date, MEDEC \cite{abacha2024medec} is one of the main datasets specifically designed for this purpose, limiting the development and evaluation of robust models. Second, the diversity of error taxonomies is often insufficient, failing to reflect the complexity of real-world medical errors. The lack of standardized expert-informed annotation protocols further results in fragmented and incomplete error representations. Third, the existing datasets focus primarily on English, limiting model development for multilingual medical contexts. Given the global nature of healthcare, the lack of multilingual resources hinders progress toward robust systems.

To address these limitations, we propose MedErrBench, a multilingual benchmark dataset for clinical error detection and correction, grounded in expert-defined error categories and supported by high-quality annotated data. An overview is provided in Figure \ref{fig: overview}. In collaboration with clinicians, we distilled potential error types into ten representative categories: Diagnosis, Management, Treatment, Pharmacotherapy, Causal Organism/Pathogen, Lab/Serum Value Interpretation, Physiology, Histology, Anatomy, and Epidemiology (see Table \ref{Tab:error_definitions} for definitions). These categories provide comprehensive coverage of clinically relevant errors and serve as a practical guideline for data annotation and system evaluation. 

Based on this typology, we construct MedErrBench in English, Arabic, and Chinese. For English and Chinese, we adapt samples from MedQA \cite{jin2021disease} by introducing expert-verified errors into clinical notes. Each instance is manually labeled with the corresponding error type. Additional erroneous examples were created by clinicians for underrepresented error types. For Arabic, we adapt samples from MedArabiQ \cite{daoud2025medarabiq}. Under the proposed typology, error types not present in the original datasets were supplemented with real-world clinical error cases contributed by practicing clinicians.
Beyond error categorization, we also label the difficulty level and reasoning type, enabling models to learn fine-grained, error-focused reasoning not supported by existing datasets.
All data across the three languages are independently reviewed by two clinicians, who corrected any issues introduced during the transformation process and validated both content accuracy and annotation quality. This rigorous pipeline ensures that MedErrBench is both clinically valid and well-suited for training and evaluating medical error detection systems.

The new proposed MedErrBench dataset supports three key clinical NLP tasks: error detection, localization, correction. To explore these tasks, we evaluate three representative classes: general-purpose LLMs, language-specific LLMs, and domain-specific medical LLMs, across English, Arabic, and Chinese. Beyond overall performance benchmarking, we conduct a series of in-depth analyses: (1) investigation of the effects of providing error-type definitions and exemplar cases;   (2) analysis of example difficulty in few-shot learning settings; (3) performance differences between knowledge-based and description-based clinical notes, and (4) cross-lingual generalization. These comprehensive experiments provide a detailed understanding of current LLM capabilities and highlight the need for clinically grounded, language-aware models in high-stakes medical applications.
Our main contributions are: 

\begin{itemize}
\item We establish a clinician-informed taxonomy of 10 clinical error types. This typology reflects real-world challenges, and provides a foundational schema for future dataset construction and evaluation in clinical NLP.

\item We introduce MedErrBench, the first fine-grained multilingual benchmark for medical error detection and correction in English, Chinese, and Arabic. It supports three tasks: error detection, localization, correction, and is rigorously validated by clinicians to ensure medical plausibility.

\item{We systematically benchmark a range of LLMs across multiple languages and model families, and conduct in-depth analyses to understand their capabilities and limitations in clinical error understanding. Our findings highlight the challenges faced by current models and motivate future research in building more robust and clinically aware systems.}

\end{itemize}

\begin{table}[t]
\centering
\caption{Classification of medical  error types, including definitions and representative examples.}
\resizebox{0.48\textwidth}{!}{
\begin{tabularx}{\textwidth}{p{2.4cm} p{5.5cm}p{6.8cm}}
\toprule
\textbf{Error Type} & \textbf{Definition} & \textbf{Example Scenario} \\
\midrule 
\textit{Diagnosis} & Failure to correctly identify the underlying condition based on clinical presentation & Interpreting myocardial infarction as GERD despite ECG abnormalities and chest pain \\ \hline
\textit{Management} & Inappropriate non-pharmacologic, non-surgical clinical decision such as observation, monitoring, or disposition & Advising “observation” in a patient with acute ST-elevation myocardial infarction \\ \hline
\textit{Treatment} & Inappropriate definitive intervention (surgical, procedural, or pharmacologic); distinct from general management & Recommending testicular biopsy in suspected torsion, delaying emergency surgery \\ \hline
\textit{Pharmacotherapy} & Incorrect drug selection, dosage, route, timing, interaction, or duration & Prescribing heparin in heparin-induced thrombocytopenia (HIT) \\ \hline
\textit{Causal Organism / Pathogen} & Misidentification of the causative microorganism in infectious disease & Attributing syphilis to \textit{Pseudomonas} instead of \textit{Treponema pallidum} \\ \hline
\textit{Lab Value Interpretation} & Misreading or misapplying diagnostic thresholds, reference ranges, or derived values & Interpreting HbA1c of 5.2\% as diagnostic for diabetes \\ \hline
\textit{Physiology} & Misconception or misinterpretation of physiological principles (e.g., ECG, PFT, etc.) & Reading an irregular rhythm without P waves as sinus rhythm rather than atrial fibrillation \\ \hline
\textit{Histology} & Misinterpretation of tissue morphology, cellular structures, or microscopic patterns & Identifying psammoma bodies from papillary thyroid carcinoma as colon adenocarcinoma features \\ \hline
\textit{Anatomy} & Errors in anatomical structure, relation, or spatial understanding & Describing the pancreas as an intraperitoneal organ \\ \hline
\textit{Epidemiology} & Misuse of statistical tools or misstatement of incidence, prevalence, or risk factors & Claiming colorectal cancer is more prevalent than breast cancer among women globally \\
\bottomrule
\end{tabularx}} 
\label{Tab:error_definitions}
\vspace{-3mm}
\end{table}

\section{Related Work}
\subsection{General Error Detection and Correction}
Error detection and correction have been applied across a range of domains, including grammatical error correction \cite{peng2025encode, ye2025excgec, Masahiro2022Interpretability}, code debugging and repair \cite{tsai2024rtlfixer, Runchu2024DebugBench}, data cleaning  \cite{reis2024generalizable}, and fact verification \cite{setty2024factcheck, Jingwei2024AFaCTA}. 
To improve performance, especially under limited annotated data, researchers have proposed techniques such as synthetic error generation \cite{Felix2024Synthetic} and auxiliary linguistic or contextual signals \cite{Yuejiao2023Enhancing}.
More recently, LLMs have been applied to error detection and correction \cite{Ryo2024Evaluating} through direct correction generation \cite{Mengsay2023Exploring} and instruction tuning \cite{Yaxin2023GrammarGPT}. Additionally, some studies leverage LLM feedback loops to refine model outputs \cite{Ryo2024When, Liangming2024Automatically, Zhibin2024CRITIC}. Although LLMs occasionally exhibit over-correction and misalignment with user intent \cite{Justin2023Closer}, human evaluations often find their corrections more fluent and acceptable compared to task-specific models \cite{Min2024Evaluating}.

\subsection{Error Detection and Correction in Healthcare}

Error detection and correction are critical in healthcare due to their impact on clinical decisions and patient safety. The MEDIQA-CORR 2024 Shared Task introduced MEDEC \cite{abacha2024medec}, the first public dataset for evaluating  errors in clinical notes based on five main error types.
Existing methods can be broadly categorized into two types: (1) prompting-based LLM strategies and (2) hybrid or traditional approaches.
Prompt-based systems employ few-shot in-context learning and chain-of-thought reasoning \cite{wu2024knowlab_aimed, gundabathula2024promptmind}, with some leveraging retrieval-augmented generation to incorporate external knowledge \cite{rajwal2024em_mixers, corbeil2024iryonlp}. Strategies include structured prompt templates, error-type hints, and self-consistency sampling. Some systems adopt in-prompt ensembling by combining outputs from multiple expert prompts, weighted by trust scores \cite{valiev2024hse}, while others rely on manual error-type categorization to guide the reasoning process. Others train models to generate rationales before proposing corrections \cite{wu2024knowlab_aimed}.
Hybrid methods combine traditional classifiers, such as support vector machines, with QA-based correction modules \cite{saeed2024medifact}. These approaches emphasize interpretability and efficiency by incorporating domain-specific features like TF-IDF scores, clinical terminology patterns, and handcrafted rules.
Nevertheless, current research predominantly depends on MEDEC, which is monolingual and lacks a broader coverage of other medical error types, constraining the generalizability and clinical applicability of proposed methods.

\section{Methodology}

\subsection{Multilingual Dataset Partitioning}
To support robust and generalizable research in medical error detection and correction, we constructed a trilingual dataset in English, Chinese, and Arabic, reflecting linguistic and regional diversity in medical education systems: English as the global scientific lingua franca, Chinese representing the world's most spoken language, and Arabic capturing the Middle East and North Africa region. We note that the datasets are not translations and are collected from multiple native-language sources, ensuring multilingual fidelity. Multi-source design introduces cross-site variability for more robust evaluation. 

The English and Chinese subsets were initially sampled from MedQA \cite{jin2021disease}, which includes clinical cases sourced from medical licensing examination questions used in the US and China. We applied filtering criteria to ensure contextual richness, removing short factoid-style questions (typically 1–2 sentences) and retaining longer, multi-sentence clinical notes that provide realistic and meaningful diagnostic or therapeutic contexts.
The Arabic subset was sampled from MedArabiQ \cite{daoud2025medarabiq} and MedAraBench \cite{Mouath2026MedAraBench}. Each Arabic question was tagged with a question style label: either scenario-based or knowledge-based. Scenario-based questions typically include a brief clinical case and require interpretation in context, whereas knowledge-based questions assess general medical facts. We used keyword heuristics (e.g., ``patient", ``age", ``child") to extract scenario-based questions, and then hand-picked more challenging knowledge-based questions with sufficient word count and specialty coverage.

\subsection{Building the Taxonomy of Clinical Error Types}

To construct a clinically grounded error typology, we collaborated with experienced clinicians to identify and refine ten representative error categories, building on MEDEC and adding five new error types: Lab/Serum Value Interpretation, Physiology, Histology, Anatomy, and Epidemiology, alongside previously established ones including Diagnosis, Management, Treatment, Pharmacotherapy, and Causal Organism/Pathogen. These categories are designed to comprehensively capture the range of factual errors commonly encountered in clinical practice and serve as a practical framework for both data annotation and system evaluation. For each error type, we provide a clear definition along with an example scenario to illustrate its typical manifestations. Table~\ref{Tab:error_definitions} provides detailed descriptions, while in the appendix Figures~\ref{fig:Error_type_En}–\ref{fig:Error_type_Ara} illustrate the ten error types in MedErrBench.

\subsection{Error Injection \& Dataset Construction}

To develop a clinically grounded dataset for medical error detection, we introduced the errors into the partitioned multilingual clinical cases.

For each question, we preserved the correct answer and randomly selected one plausible but incorrect alternative, while discarding the remaining distractors. Using these selected answers, we constructed two versions of a clinical note: one in which the correct answer was naturally integrated into the context, and another in which the incorrect answer was inserted in its place. 

The original datasets lacked certain error types. To address this gap and ensure comprehensive coverage, we collaborated with experienced clinicians who contributed real-world clinical cases for the missing categories. Specifically, the English dataset originally lacked Physiology, Histology, Anatomy, and Epidemiology examples, while the Arabic dataset lacked Lab/Serum Value Interpretation examples. To operationalize this construction process, we employed LLMs to assist in transforming the original samples into full-length clinical narratives in all three languages. Supplementary Figures \ref{fig: Prompt_EN}–\ref{fig: Prompt_Ara} provide the prompts for reproducibility, while Figures~\ref{fig:Example_EN}-\ref{fig:Example_Ara} illustrate error insertion examples from MedErrBench.

\subsection{Important Clinical Words, Difficulty-Level and Reasoning-Type Annotation }

To better analyze task complexity and enable more fine-grained evaluation of model capabilities, we additionally manually annotate each instance with three auxiliary attributes: important clinical words, difficulty level, and reasoning type. Important clinical words capture the most salient concept or decision point in each case (e.g., a diagnosis, therapeutic action, or critical finding), highlighting the key linguistic cues that both humans and models must attend to during error detection.
Each instance was assigned one of three difficulty levels: Easy, Medium, or Hard. The annotation process followed a set of predefined clinical reasoning guidelines that considered multiple factors, including the clarity of clinical cues, the rarity or complexity of the underlying condition, the number of reasoning steps required to reach a correct conclusion, and the overall length and ambiguity of the question text.
Additionally, each item was annotated according to the type of reasoning required to answer it, using a three-level classification scheme: Factual Recall, Single-hop Reasoning, and Multi-hop Reasoning. These categories reflect increasing levels of inferential complexity and are critical for evaluating the diagnostic reasoning abilities of LLMs.

\subsection{Expert Review and Quality Control}

We employed a rigorous two-stage human review and quality control process. In the first stage, three native-speaking NLP researchers (English, Chinese, and Arabic), after studying the clinician-defined taxonomy of ten clinical error types, performed initial annotation and manual verification. This process involved identifying hallucinated or unnatural content, removing incorrect or extraneous information introduced by LLMs, and segmenting overly long sentences, particularly in the Chinese dataset, where punctuation such as commas often failed to separate clauses properly.

In the second stage, two independent clinicians reviewed all instances to ensure medical validity. They corrected transformation errors, validated error labels, and verified key clinical terms, difficulty levels, and reasoning types. Disagreements were categorized as logical errors, misclassifications, or typos; typos were corrected directly, while other issues were revised if flagged by either clinician. No conflicting corrections were proposed.

\section{Experiment and Results}

\subsection{Evaluation Metrics}
We evaluated model performance across three sub-tasks: error detection, error localization, sentence correction. We use accuracy, ROUGE \cite{lin2004rouge} (including ROUGE-1, ROUGE-2, ROUGE-L), BLEU \cite{Kishore2002Bleu}, BERTScore \cite{Tianyi2020BERTScore}, and BLEURT \cite{Thibault2020BLEURT} as the main evaluation metrics\footnote{Due to page limitations, we report Accuracy, ROUGE-1, BERTScore, and BLEURT in the main paper, and present the remaining evaluation results in the appendix.}.

\subsection{Baseline Models}
We evaluate a diverse set of recent language models, grouped by their design objectives:
\\ \textbf{Group 1: General-purpose LLMs.} This group includes models developed for broad, cross-domain language tasks. We evaluate GPT-4o \cite{openai2024gpt4o}, GPT-4o-mini \cite{openai2024gpt4omini}, Gemini 2.5 Flash Lite \cite{google2025medgemma}, Gemini 2.0 Flash \cite{google2025medgemma}, LLaMA3-8B \cite{meta2024llama3} and Llama-3.3-70B-Instruct\footnote{https://huggingface.co/meta-llama/Llama-3.3-70B-Instruct}.
\\ \textbf{Group 2: Language-specialized LLMs.} These models are primarily optimized for specific languages. Our selection includes Chinese models (
Qwen2.5-7B-Instruct \cite{qwen2024}, DeepSeek-R1 \cite{deepseekr12024}, DeepSeek-V3 \cite{deepseekv32024}, Doubao-1.5-Thinking-Pro \cite{doubao2024} and Arabic model\footnote{We evaluated Falcon and Jais; however, both exhibited issues for error detection and correction task. For example, Falcon-Arabic consistently returned \texttt{<text id> 0 -1 NA} for all cases. Therefore, we do not report results for these two models.} ALLAM-7B \cite{bari2025allam}.
\\ \textbf{Group 3: Medical-domain LLMs.} This group comprises models specifically designed for clinical and biomedical applications. We include MedGemma-4B \cite{google2025medgemma, medgemma2025blog}, MedGemma-27B \cite{google2025medgemma, medgemma2025blog}, and HuatuoGPT-o1-7B \cite{huatuogpt2024}.

\begin{figure}[t]
     \centering
\includegraphics[width=0.45\textwidth]{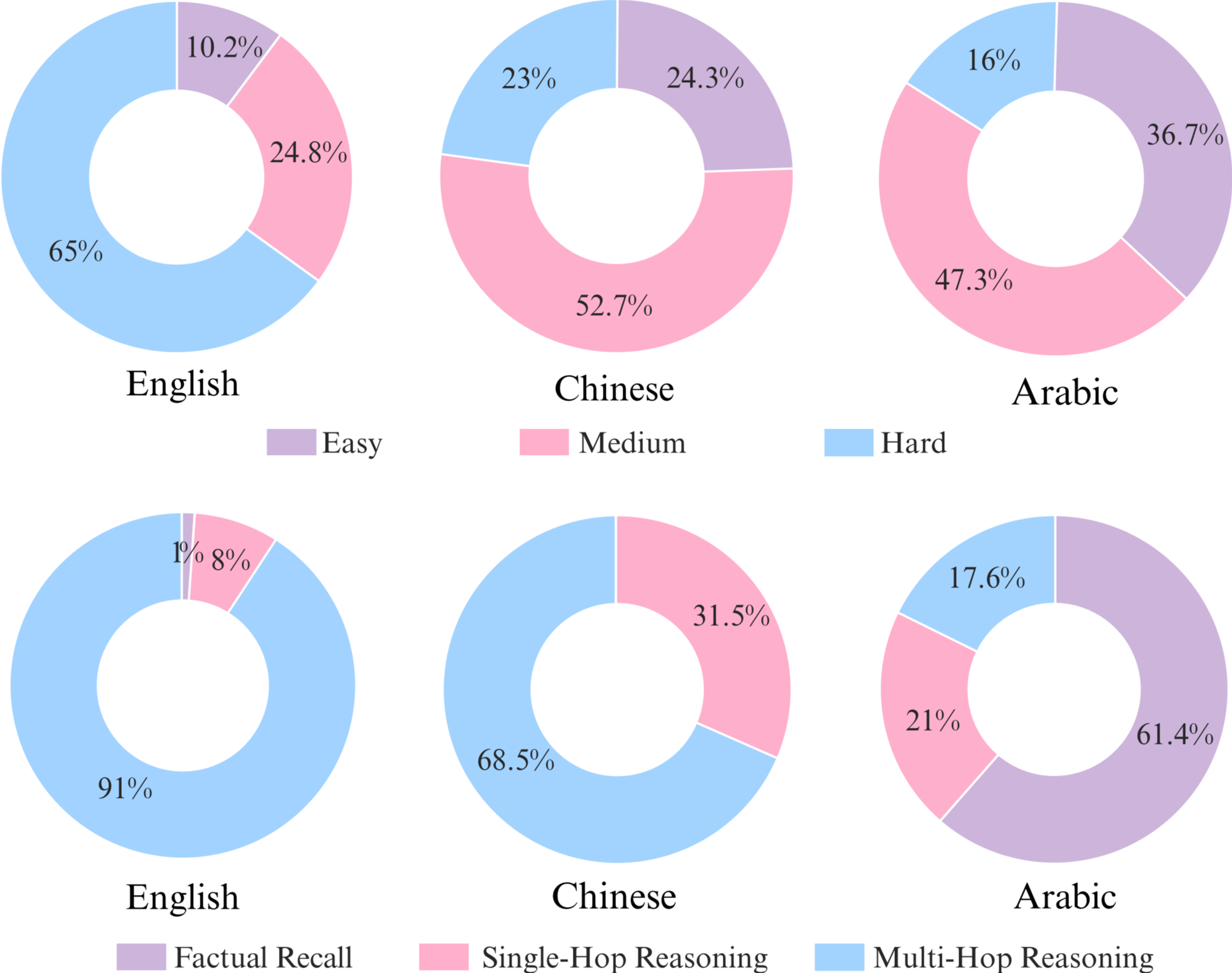}
    \caption{Distribution of difficulty level and reasoning type.}
    \label{fig: Difficulty}
    \vspace{-3mm}
\end{figure}

\subsection{Overview of Dataset}

Figure \ref{fig: Difficulty} shows the distribution of difficulty level for all three languages. The English dataset is skewed toward higher-difficulty content, with 65\% of questions labeled as Hard, reflecting a greater emphasis on abstraction and multi-step integration. The Chinese dataset shows a more balanced distribution, with 52.7\% of questions categorized as Medium, and roughly equal proportions of Easy (24.3\%) and Hard (23\%) items. In contrast, the Arabic dataset contains a higher proportion of Medium questions (47.3\%) and fewer Hard items (16\%), suggesting simpler question formats and a stronger emphasis on factual recall. 

Figure \ref{fig: Difficulty} also shows the distribution of reasoning type for all three languages. The English dataset is overwhelmingly multi-hop in nature, with 91\% of questions requiring the integration of multiple clinical elements, consistent with case-based diagnostic reasoning. Chinese questions show a more moderate distribution, with 68.5\% classified as Multi-hop and 31.5\% as Single-hop, indicating a balance between direct inference and integrative tasks. Arabic questions, by contrast, are predominantly based on Factual Recall (61.4\%), with lower proportions of Single-hop (21\%) and Multi-hop (17.6\%) reasoning. This variation in reasoning types across datasets aligns closely with the observed difficulty distributions and reinforces the need for language-specific modeling and evaluation strategies.

In the English dataset, 7.5\% of the test set instances were found to be mislabeled. The most frequent issue involved confusion between the Management and Treatment categories, accounting for approximately 60\% of all labeling errors. Additional misclassifications included examples such as Diagnosis incorrectly labeled as Management, as well as errors involving Pharmacotherapy, Epidemiology, and Lab/Serum Value Interpretation. In the Chinese dataset, two clinicians independently reviewed the data and identified issues in 5\% and 6.5\% of instances, respectively. The most common concern was the expression of clinical measurement units, which often did not align with standard clinical usage. 
In the Arabic subset, clinical review identified issues in approximately 12\% of the data. These included outdated clinical procedures, corrections that were inapplicable or clinically inappropriate, and other medical inaccuracies. Additionally, clinicians flagged a further 5\% of the entries for spelling mistakes. Eight error-type misclassifications were also identified, including three physiology-related questions that had been incorrectly categorized under other types. Detailed data splits, basic statistics, and the distribution of error types are provided in Table \ref{tab: statistics} and Figure \ref{fig: Error_Distribution} in the appendix.

\renewcommand{\arraystretch}{0.75}
\begin{table}[ht]
\centering
\caption{Results on MedErrBench-EN.}
\vspace{-1mm}
\label{tab:clinmed-english-results}
\resizebox{0.48\textwidth}{!}{%
\begin{tabular}{lccccc}
\toprule
\textbf{Models} & \multicolumn{1}{c}{\textbf{Detection}} & \multicolumn{1}{c}{\textbf{Localization}} & \multicolumn{3}{c}{\textbf{Error Correction}} \\
\textbf{} & \textbf{Accuracy} & \textbf{Accuracy} & \textbf{ROUGE-1} & \textbf{BertScore} & \textbf{BLEURT} \\
\midrule
\rowcolor{gray!15} \multicolumn{6}{c}{\textit{General-purpose LLMs}} \\
\midrule
gpt-4o                  & 0.596 & 0.346 & 0.415 & 0.428 & 0.407 \\
gpt-4o-mini             & \underline{0.664} & 0.524 & 0.487 & 0.498 & 0.472 \\
Gemini 2.5 Flash Lite   & 0.567 & 0.264 & 0.349 & 0.362 & 0.346 \\
Gemini 2.0 Flash        & 0.514 & 0.168 & 0.281 & 0.294 & 0.288 \\
Llama3-8b               & 0.519 & 0.361 & 0.266 & 0.261 & 0.282 \\
Llama-3.3-70B-Instruct  & 0.582 & 0.255 & 0.369 & 0.369 & 0.385 \\
\midrule
\rowcolor{gray!15} \multicolumn{6}{c}{\textit{Language-specialized LLMs}} \\
\midrule
Qwen2.5-7B-Instruct     & 0.563 & 0.490 & 0.372 & 0.450 & 0.371 \\
Deepseek-R1             & 0.582 & 0.577 & 0.700 & 0.716 & 0.681 \\
Deepseek-V3             & 0.587 & \underline{0.582} & \underline{0.703} & \underline{0.732} & \underline{0.693} \\
Doubao-1.5              & \textbf{0.779} & \textbf{0.774} & \textbf{0.766} & \textbf{0.783} & \textbf{0.773} \\
ALLAM-7B                & 0.029 & 0.014 & 0.015 & 0.020 & 0.014 \\
\midrule
\rowcolor{gray!15} \multicolumn{6}{c}{\textit{Medical-domain LLMs}} \\
\midrule
MedGemma-4b       & 0.505 & 0.438 & 0.511 & 0.518 & 0.513 \\
MedGemma-27b      & 0.543 & 0.245 & 0.377 & 0.390 & 0.349 \\
HuatuoGPT-o1-7b   & 0.574 & 0.530 & 0.486 & 0.475 & 0.475 \\
\bottomrule
\end{tabular}%
\vspace{-1mm}
}
\end{table}

\renewcommand{\arraystretch}{0.85}
\begin{table}[ht]
\centering
\caption{Results on MedErrBench-CN.}
\vspace{-1mm}
\label{tab:clinmed-chinese-results}
\resizebox{0.48\textwidth}{!}{%
\begin{tabular}{lccccc}
\toprule
\textbf{Models} & \multicolumn{1}{c}{\textbf{Detection}} & \multicolumn{1}{c}{\textbf{Localization}} & \multicolumn{3}{c}{\textbf{Error Correction}} \\
\textbf{} & \textbf{Accuracy} & \textbf{Accuracy} & \textbf{ROUGE-1} & \textbf{BertScore} & \textbf{BLEURT} \\
\midrule
\rowcolor{gray!15} \multicolumn{6}{c}{\textit{General-purpose LLMs}} \\
\midrule
gpt-4o                     & 0.630 & 0.205 & 0.265 & 0.365 & 0.266 \\
gpt-4o-mini               & 0.505 & 0.115 & 0.244 & 0.390 & 0.257 \\
Gemini 2.5 Flash Lite     & 0.600 & 0.375 & 0.448 & 0.533 & 0.455 \\
Gemini 2.0 Flash          & 0.705 & 0.455 & 0.569 & 0.659 & 0.577 \\
Llama3-8b                 & 0.500 & 0.320 & 0.416 & 0.532 & 0.483 \\
Llama-3.3-70B-Instruct	 & 0.675 & 0.380 & 0.506 & 0.606 & 0.509 \\
\midrule
\rowcolor{gray!15} \multicolumn{6}{c}{\textit{Language-specialized LLMs}} \\
\midrule
Qwen2.5-7B-Instruct    & 0.625 & 0.570 & 0.493 & 0.576 & 0.462 \\
Deepseek-R1            & \underline{0.735} & \underline{0.705} & \underline{0.802} & \underline{0.851} & \underline{0.781} \\
Deepseek-V3            & 0.650 & 0.640 & \textbf{0.833} & \textbf{0.873} & \textbf{0.806} \\
Doubao-1.5             & \textbf{0.750} & \textbf{0.735} & 0.788 & 0.835 & 0.777 \\
ALLAM-7B               & 0.395 & 0.340 & 0.284 & 0.360 & 0.286 \\
\midrule
\rowcolor{gray!15} \multicolumn{6}{c}{\textit{Medical-domain LLMs}} \\
\midrule
MedGemma-4b            & 0.525 & 0.500 & 0.549 & 0.581 & 0.547 \\
MedGemma-27b           & 0.605 & 0.285 & 0.441 & 0.537 & 0.438 \\
HuatuoGPT-o1-7b        & 0.525 & 0.275 & 0.167 & 0.545 & 0.530 \\
\bottomrule
\end{tabular}%
}
\vspace{-1mm}
\end{table}

\renewcommand{\arraystretch}{0.85}
\begin{table}[ht]
\centering
\caption{Results on MedErrBench-Ara. }
\vspace{-1mm}
\label{tab:clinmed-arabic-results}
\resizebox{0.48\textwidth}{!}{%
\begin{tabular}{lccccc}
\toprule
\textbf{Models} & \multicolumn{1}{c}{\textbf{Detection}} & \multicolumn{1}{c}{\textbf{Localization}} & \multicolumn{3}{c}{\textbf{Error Correction}} \\
\textbf{} & \textbf{Accuracy} & \textbf{Accuracy} & \textbf{ROUGE-1} & \textbf{BertScore} & \textbf{BLEURT} \\
\midrule
\rowcolor{gray!15} \multicolumn{6}{c}{\textit{General-purpose LLMs}} \\
\midrule
gpt-4o                & \underline{0.680} & 0.320 & 0.399 & 0.592 & 0.414 \\
gpt-4o-mini           & 0.577 & 0.175 & 0.260 & 0.469 & 0.292 \\
Gemini 2.5 Flash Lite & 0.495 & 0.268 & 0.303 & 0.432 & 0.318 \\
Gemini 2.0 Flash      & 0.598 & 0.299 & 0.315 & 0.503 & 0.332 \\
Llama3-8b             & 0.371 & 0.309 & 0.311 & 0.324 & 0.313 \\
Llama-3.3-70B-Instruct	&0.557	&0.381	&0.412	&0.454	&0.405 \\
\midrule
\rowcolor{gray!15} \multicolumn{6}{c}{\textit{Language-specialized LLMs}} \\
\midrule
Qwen2.5-7B-Instruct            & 0.536 & 0.381 & 0.329 & 0.473 & 0.353 \\
Deepseek-R1            & \textbf{0.711} & \textbf{0.505} & 0.568 & \underline{0.756} & \underline{0.610} \\
Deepseek-V3            & 0.608 & \textbf{0.505} & \textbf{0.677} & \textbf{0.814} & \textbf{0.699} \\
Doubao-1.5             & 0.670 & \textbf{0.505} & \underline{0.582} & 0.736 & 0.583 \\
ALLAM-7B                 & 0.072 & 0.021 & 0.045 & 0.049 & 0.046 \\
\midrule
\rowcolor{gray!15} \multicolumn{6}{c}{\textit{Medical-domain LLMs}} \\
\midrule
MedGemma-4b       & 0.454 & 0.433 & 0.438 & 0.450 & 0.439 \\
MedGemma-27b      & 0.552 & 0.240 & 0.266 & 0.456 & 0.286 \\
HuatuoGPT-o1-7b   & 0.371 & 0.397 & 0.302 & 0.450 & 0.420 \\
\bottomrule
\end{tabular}%
\vspace{-1mm}
}
\end{table}

\begin{figure*}[ht]
    \centering
\includegraphics[width=0.7\textwidth]{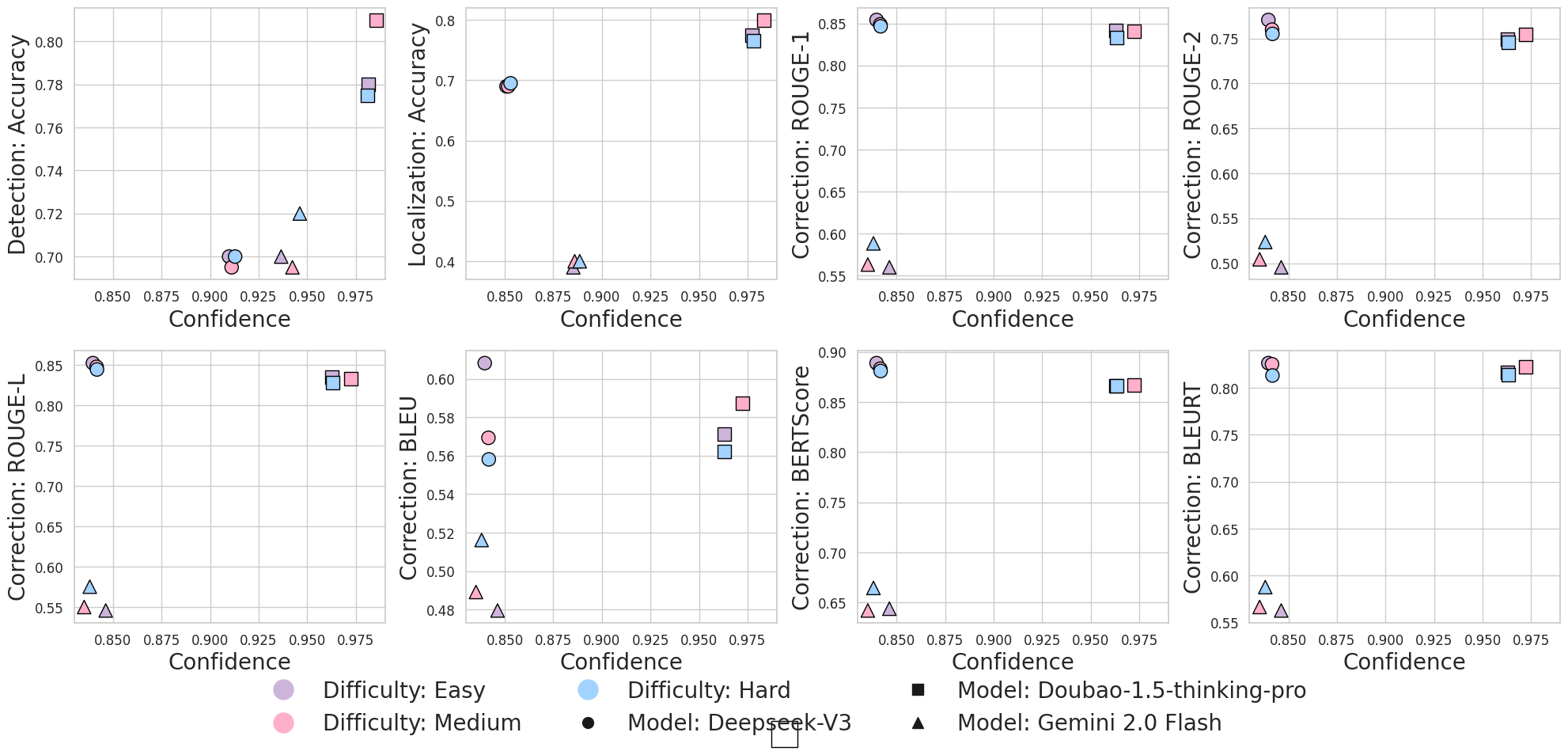}
    \vspace{-2mm}
    \caption{Performance comparison across easy, medium, and hard examples in few-shot learning.} 
    \vspace{-6mm}
    \label{fig: scatter_plot_diffculty}
\end{figure*}

\subsection{Overall Performance}

We evaluated the models on three core tasks across the English, Chinese, and Arabic datasets (Tables \ref{tab:clinmed-english-results}-\ref{tab:clinmed-arabic-results}). Detailed results for all evaluation metrics are reported in Tables \ref{tab:full-clinmed-english-results}-\ref{tab:full-clinmed-arabic-results} in the appendix.
Overall, Doubao-1.5-thinking-pro, Deepseek-R1, Deepseek-V3 perform better than other models across the languages. 
Despite being trained on domain-specific data, medical LLMs like MedGemma and HuatuoGPT do not consistently outperform general-purpose models. MedGemma models are trained on medical text, medical QA, EHR, and medical images, while HuatuoGPT leverages medical exam questions. These models focus on domain knowledge and factual recall rather than error detection and correction, which requires broader linguistic reasoning and robustness to noisy clinical text.
The Arabic LLM underperformed significantly perhaps since it lacks medical domain adaptation. We tested three different prompts for ALLAM-7B, yielding an average detection accuracy of only 0.083, suggesting that prompt sensitivity alone does not explain the failure.

The drop in accuracy from detection to localization is expected, as detection is a binary task, whereas localization is a more complex multi-class task requiring precise error-span identification; localization failures typically arise when an error is detected but its span is misidentified. Overall, localization and correction remain more challenging than detection across all models and languages, as reflected by their consistently lower scores. We did not observe any error type that consistently challenged all models. However, Llama3-8B showed weaker performance on Management errors in Arabic, while ALLAM-7B and Qwen2.5-7B-Instruct struggled primarily with Anatomy, Causal Organism/Pathogen, and Diagnosis errors.
Llama3-8B exhibits signs of overcorrection in Arabic: frequent NA predictions, combined with the metric’s assignment of a score of 1 when both predicted and true labels are NA, artificially inflate performance by increasing true negatives.
Meanwhile, models like ALLAM-7B fail on out-of-domain or cross-lingual tasks, highlighting the importance of robust multilingual evaluation. 
A clinician performed a human evaluation of 50 Chinese samples, yielding average scores of 0.52 for Gemini 2.0 Flash and 0.17 for GPT-4o-mini.

\renewcommand{\arraystretch}{0.85}
\begin{table}[t]
\centering
\caption{Performance comparison of models under different error type Conditions. ``ET"  and ``DEF" indicate error types and definitions, respectively.}
\label{tab:error-type-analysis}
\resizebox{0.48\textwidth}{!}{%
\begin{tabular}{lcccccc}
\toprule
\textbf{} & \multicolumn{1}{c}{\textbf{Detection}} & \multicolumn{1}{c}{\textbf{Localization}} & \multicolumn{3}{c}{\textbf{Error Correction}} \\
\textbf{} & \textbf{Accuracy} & \textbf{Accuracy} & \textbf{ROUGE-1} & \textbf{BertScore} & \textbf{BLEURT} \\
\midrule
\rowcolor{gray!15} \multicolumn{6}{c}{\textit{Deepseek-V3 (Zero-shot)}} \\
\midrule
\textit{w/o ET \& DEF}
      & 0.690 & 0.645 & 0.660 & 0.735 & 0.599 \\
\textit{w/o DEF}
      & 0.625 & 0.610 & 0.731 & 0.795 & 0.685 \\
\textit{w ET \& DEF}
      & 0.650 & 0.640 & 0.730 & 0.794 & 0.684 \\
\midrule
\rowcolor{gray!15} \multicolumn{6}{c}{\textit{Deepseek-V3 (Few-shot)}} \\
\midrule
\textit{w/o ET \& DEF}
      & 0.720 & 0.690 & 0.695 & 0.767 & 0.639 \\
\textit{w/o DEF}
      & 0.710 & 0.705 & 0.736 & 0.796 & 0.684 \\
\textit{w ET \& DEF}
      & 0.715 & 0.715 & 0.763 & 0.821 & 0.705 \\
\midrule
\rowcolor{gray!15} \multicolumn{6}{c}{\textit{Doubao-1.5-thinking-pro (Zero-shot)}} \\
\midrule
\textit{w/o ET \& DEF}
      & 0.695 & 0.640 & 0.637 & 0.727 & 0.607 \\
\textit{w/o DEF}
      & 0.730 & 0.710 & 0.673 & 0.751 & 0.646 \\
\textit{w ET \& DEF}
      & 0.750 & 0.725 & 0.669 & 0.728 & 0.636 \\
\midrule
\rowcolor{gray!15} \multicolumn{6}{c}{\textit{Doubao-1.5-thinking-pro (Few-shot)}} \\
\midrule
\textit{w/o ET \& DEF}
      & 0.735 & 0.695 & 0.707 & 0.765 & 0.651 \\
\textit{w/o DEF}
      & 0.765 & 0.750 & 0.729 & 0.777 & 0.671 \\
\textit{w ET \& DEF}
      & 0.775 & 0.765 & 0.699 & 0.753 & 0.665 \\
\bottomrule
\end{tabular}%
}
\end{table}

\subsection{Impact of Providing Examples and Error Type Definitions}

Table \ref{tab:error-type-analysis} presents the performance of models under varying configurations of Error Type (ET) and Definition (DEF) availability. Detailed evaluation results are provided in Table~\ref{tab:full-error-type-analysis} in the appendix. Overall, we observe that providing error type definitions consistently improves performance, particularly in the zero-shot setting. For example, in the Deepseek-V3 (Zero-shot) setup, adding definitions (\textit{w DEF} or \textit{w ET \& DEF}) boosts error correction metrics compared to the baseline (\textit{w/o ET \& DEF}). This highlights that semantic clarity from definitions is beneficial even without structural labels like error types.

Additionally, few-shot configurations consistently outperform their zero-shot counterparts across all models and settings, indicating that in-context examples provide strong guidance for both detection and correction tasks. 
Interestingly, while providing both ET and DEF is generally helpful, the isolated impact of error types alone (i.e., \textit{w/o DEF}) can be inconsistent. In some few-shot settings (e.g., Doubao-1.5-thinking-pro), adding ETs without definitions slightly improves detection but may reduce correction performance, suggesting potential cognitive overload or prompt misalignment.

The performance divergence between Deepseek-V3 and Doubao under different prompt settings may stem from their architectural and training differences. Deepseek-V3 appears more responsive to structured definitions and error-type annotations, possibly due to multilingual and multitask training, which enhances its generalization across abstract prompt forms. In contrast, Doubao demonstrates stronger alignment with example-driven prompts, but exhibits sensitivity or degradation when additional structured elements (e.g., both ET and DEF) are introduced, especially in few-shot scenarios. This suggests that prompt design must consider model-specific alignment and interpretability characteristics, as misaligned guidance may counterintuitively hinder performance.

\begin{figure*}[ht]
    \centering
\includegraphics[width=0.65\textwidth]{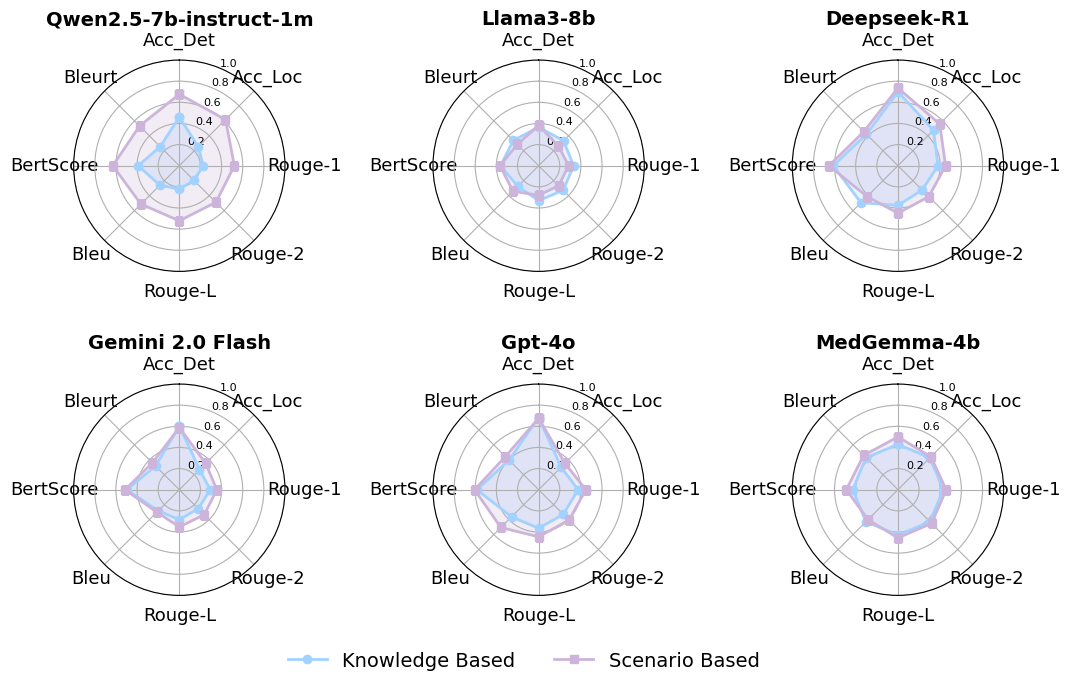}
    \vspace{-2mm}
    \caption{Comparison of models based on knowledge-based and description-based evaluation.} 
    \label{fig: SvsK}
    \vspace{-3mm}
\end{figure*}

\subsection{Impact of Providing Example Difficulty Levels in Few-shot Learning Settings}

Figure \ref{fig: scatter_plot_diffculty} illustrates the relationship between example difficulty levels and model performance in few-shot learning across three baseline models with good and stable performance on Chinese dataset: Deepseek-V3 (circle), Doubao-1.5-thinking-pro (square), and Gemini 2.0 Flash (triangle).
For Doubao-1.5-thinking-pro, the performance trend across difficulty levels follows a consistent pattern of Medium $>$ Easy $>$ Hard across nearly all metrics. 
In contrast, Deepseek-V3 shows a different behavior. For correction tasks, the model clearly follows Easy $>$ Medium $>$ Hard. For localization, Deepseek-V3 performs similarly across all difficulty levels, with Hard examples even slightly outperforming the others.
Gemini 2.0 Flash, however, displays an opposite trend to Doubao-1.5-thinking-pro and Deepseek-V3. For most metrics, the order is Hard $>$ Medium $>$ Easy, with overall lower performance. This may imply that the task is relatively more difficult for Gemini 2.0 Flash, and that harder in-context examples are more informative and helpful for its learning. 
In terms of confidence scores, Doubao-1.5-thinking-pro exhibits the highest confidence across all difficulty levels, noticeably exceeding both Deepseek-V3 and Gemini 2.0 Flash. This suggests that Doubao-1.5-thinking-pro is more certain in its predictions, although confidence does not always correlate perfectly with correctness, especially in challenging scenarios.

\subsection{Analysis of Knowledge vs. Scenario-based Data}

Figure \ref{fig: SvsK} shows the  evaluation results of six LLMs on an Arabic dataset, segmented by knowledge-based and scenario-based clinical notes. The goal is to assess whether LLMs can capture language-independent conceptions. Most models perform better on scenario-based tasks, suggesting a reliance on contextual pattern recognition rather than robust internal medical knowledge in Arabic.
An exception is Llama3-8b, which slightly performs better on knowledge-based tasks. This may indicate that its learned representations are more tightly coupled with language-independent factual knowledge, allowing it to resist some misconceptions when direct medical facts are queried. 
Qwen2.5-7B-Instruct exhibits the largest performance gap which suggests that the model is over-reliant on surface-level patterns and instruction-following heuristics, making it more vulnerable to reproducing misconceptions in structured factual queries, especially in low-resource languages like Arabic. We also evaluate the cross-lingual generalization, please refer to the appendix section \ref {appendix: Cross-lingual}.

\section{Conclusion and Discussion}
In this work, we present a novel multilingual benchmark dataset for clinical error detection and correction, grounded in expert-defined error categories and validated across English, Chinese, and Arabic. The work addresses key limitations in existing resources by providing diverse, high-quality annotations that reflect the complexity of real-world medical errors in multiple languages. Through extensive evaluation of various LLMs, we reveal the current challenges in automated clinical error understanding and emphasize the importance of clinically informed and language-specific approaches. Our dataset and analysis lay a solid foundation for advancing research in clinical NLP focused on improving patient safety. In addition to detection, localization, and correction, our dataset supports tasks such as error classification, key concept extraction, difficulty assessment, and reasoning type classification, enabling new avenues for fine-grained clinical reasoning in NLP models.
Our dataset in this component focuses on errors in professional medical knowledge. In future work, we plan to expand it with real clinical data to further enhance its coverage and completeness. Our future efforts will also focus on (1) increasing the overall scale and diversity of the corpus; (2) developing multi-agent systems to improve LLM performance on clinical tasks; (3) advancing evaluation methodologies for medical error detection and correction; (4) integrating severity stratification and assessing harm-reduction strategies.

\section*{Limitations}

We note two limitations of the present study. First, the proposed dataset does not include explicit annotations for severity levels or equity-related dimensions. Our primary objective is to establish a robust multilingual foundation for medical error detection and correction, rather than to exhaustively characterize downstream clinical risk or fairness properties. In this sense, the dataset represents an initial step toward such analyses, and is, to our knowledge, only the second publicly available resource after MEDEC that addresses medical errors in a multilingual setting. Future work can build upon this foundation by incorporating severity stratification and equity-aware annotations.
Second, the Arabic portion of the dataset remains under active expansion. Due to the limited availability of high-quality  publicly accessible medical data in this low-resource language, the current Arabic subset is smaller and less diverse than those of higher-resource languages. We plan to continue data collection and curation efforts to further expand and balance the dataset across languages, which may improve both coverage and robustness in future iterations.

\section*{Acknowledgments}
This work was supported by the Meem Foundation, the NYUAD Center for Artificial Intelligence and Robotics, funded by Tamkeen under the NYUAD Research Institute Award CG010, the Center for Brain and Health, funded by Tamkeen under NYU Abu Dhabi Research Institute grant CG012, and the Center for Interdisciplinary Data Science \& AI (CIDSAI), funded by Tamkeen under the NYUAD Research Institute Award CG016. The research was carried out on the High Performance Computing resources at New York University Abu Dhabi (Jubail). We are grateful to Dr. Xinwei Hou and Dr. Zijie Fang for their support and contributions to data annotation.



\bibliography{mybib}

@article{Mouath2026MedAraBench,
  title={MedAraBench: Large-Scale Arabic Medical Question Answering Dataset and Benchmark},
  author={Mouath, Abu-Daoud and Leen, Kharouf and Omar, El Hajj and Dana, El Samad and Mariam,  Al-Omari and Jihad, Mallat and Khaled, Saleh and Nizar, Habash and Farah, E. Shamout},
  journal={arXiv preprint arXiv:2602.01714},
  year={2026}
}

@inproceedings{pajaro2024verbanexai,
  title={Verbanexai at mediqa-corr: Efficacy of gru with biowordvec and clinicalbert in error correction in clinical notes},
  author={Pajaro, Juan and Puertas, Edwin and Villate, David and Estrada, Laura and Tinjaca, Laura},
  booktitle={Proceedings of the 6th clinical natural language processing workshop},
  pages={461--469},
  year={2024}
}

@inproceedings{jadhav2024maven,
  title={Maven at mediqa-corr 2024: Leveraging rag and medical llm for error detection and correction in medical notes},
  author={Jadhav, Suramya and Shanbhag, Abhay and Joshi, Sumedh and Date, Atharva and Sonawane, Sheetal},
  booktitle={Proceedings of the 6th Clinical Natural Language Processing Workshop},
  pages={374--381},
  year={2024}
}

@inproceedings{alzghoul2024cld,
  title={Cld-mec at mediqa-corr 2024 task: Gpt-4 multi-stage clinical chain of thought prompting for medical errors detection and correction},
  author={Alzghoul, Renad and Ayaabdelhaq, Ayaabdelhaq and Tabaza, Abdulrahman and Altamimi, Ahmad},
  booktitle={Proceedings of the 6th Clinical Natural Language Processing Workshop},
  pages={537--556},
  year={2024}
}

@inproceedings{Kishore2002Bleu,
  author       = {Kishore Papineni and
                  Salim Roukos and
                  Todd Ward and
                  Wei{-}Jing Zhu},
  title        = {Bleu: a Method for Automatic Evaluation of Machine Translation},
  booktitle    = {Proceedings of the 40th Annual Meeting of the Association for Computational
                  Linguistics, July 6-12, 2002, Philadelphia, PA, {USA}},
  pages        = {311--318},
  year         = {2002}
}

@incollection{soori2024errors,
  title={Errors in medical procedures},
  author={Soori, Hamid},
  booktitle={Errors in Medical Science Investigations},
  pages={205--224},
  year={2024}
}

@article{newman2021rate,
  title={Rate of diagnostic errors and serious misdiagnosis-related harms for major vascular events, infections, and cancers: toward a national incidence estimate using the “Big Three”},
  author={Newman-Toker, David E and Wang, Zheyu and Zhu, Yuxin and Nassery, Najlla and Tehrani, Ali S Saber and Schaffer, Adam C and Yu-Moe, Chihwen Winnie and Clemens, Gwendolyn D and Fanai, Mehdi and Siegal, Dana},
  journal={Diagnosis},
  volume={8},
  number={1},
  pages={67--84},
  year={2021}
}

@article{ahsani2022interventions,
  title={Interventions to reduce the incidence of medical error and its financial burden in health care systems: A systematic review of systematic reviews},
  author={Ahsani-Estahbanati, Ehsan and Sergeevich Gordeev, Vladimir and Doshmangir, Leila},
  journal={Frontiers in medicine},
  volume={9},
  pages={875426},
  year={2022}
}

@article{anjum2024patient,
  title={Patient Safety and Quality Improvement: Reducing Medical Errors in Healthcare},
  author={Anjum, Fauzia and Din, Brigadier Raffi Ud and Ashraf, Saira},
  journal={Multidisciplinary Journal of Healthcare (MJH)},
  volume={1},
  number={2},
  pages={13--23},
  year={2024}
}

@inproceedings{Jingwei2024AFaCTA,
  author       = {Jingwei Ni and
                  Minjing Shi and
                  Dominik Stammbach and
                  Mrinmaya Sachan and
                  Elliott Ash and
                  Markus Leippold},
  title        = {AFaCTA: Assisting the Annotation of Factual Claim Detection with Reliable
                  {LLM} Annotators},
  booktitle    = {Proceedings of the 62nd Annual Meeting of the Association for Computational
                  Linguistics (Volume 1: Long Papers), {ACL} 2024, Bangkok, Thailand,
                  August 11-16, 2024},
  pages        = {1890--1912},
  year         = {2024}
}

@inproceedings{setty2024factcheck,
  title={Factcheck editor: Multilingual text editor with end-to-end fact-checking},
  author={Setty, Vinay},
  booktitle={Proceedings of the 47th International ACM SIGIR Conference on Research and Development in Information Retrieval},
  pages={2744--2748},
  year={2024}
}

@article{reis2024generalizable,
  title={Generalizable Data Cleaning of Tabular Data in Latent Space},
  author={Reis, Eduardo and Abdelaal, Mohamed and Binnig, Carsten},
  journal={Proceedings of the VLDB Endowment},
  volume={17},
  number={13},
  pages={4786--4798},
  year={2024}
}

@inproceedings{Runchu2024DebugBench,
  author       = {Runchu Tian and
                  Yining Ye and
                  Yujia Qin and
                  Xin Cong and
                  Yankai Lin and
                  Yinxu Pan and
                  Yesai Wu and
                  Haotian Hui and
                  Weichuan Liu and
                  Zhiyuan Liu and
                  Maosong Sun},
  title        = {DebugBench: Evaluating Debugging Capability of Large Language Models},
  booktitle    = {Findings of the Association for Computational Linguistics, {ACL} 2024,
                  Bangkok, Thailand and virtual meeting, August 11-16, 2024},
  pages        = {4173--4198},
  year         = {2024}
}

@inproceedings{tsai2024rtlfixer,
  title={Rtlfixer: Automatically fixing rtl syntax errors with large language model},
  author={Tsai, YunDa and Liu, Mingjie and Ren, Haoxing},
  booktitle={Proceedings of the 61st ACM/IEEE Design Automation Conference},
  pages={1--6},
  year={2024}
}

@inproceedings{Yuejiao2023Enhancing,
  author       = {Yuejiao Fei and
                  Leyang Cui and
                  Sen Yang and
                  Wai Lam and
                  Zhenzhong Lan and
                  Shuming Shi},
  title        = {Enhancing Grammatical Error Correction Systems with Explanations},
  booktitle    = {Proceedings of the 61st Annual Meeting of the Association for Computational
                  Linguistics (Volume 1: Long Papers), {ACL} 2023, Toronto, Canada,
                  July 9-14, 2023},
  pages        = {7489--7501},
  year         = {2023}
}

@inproceedings{Felix2024Synthetic,
  author       = {Felix Stahlberg and
                  Shankar Kumar},
  title        = {Synthetic Data Generation for Low-resource Grammatical Error Correction
                  with Tagged Corruption Models},
  booktitle    = {Proceedings of the 19th Workshop on Innovative Use of {NLP} for Building
                  Educational Applications, {BEA} 2024, Mexico City, Mexico, June 20,
                  2024},
  pages        = {11--16},
  year         = {2024}
}

@inproceedings{Min2024Evaluating,
  author       = {Min Zeng and
                  Jiexin Kuang and
                  Mengyang Qiu and
                  Jayoung Song and
                  Jungyeul Park},
  title        = {Evaluating Prompting Strategies for Grammatical Error Correction Based
                  on Language Proficiency},
  booktitle    = {Proceedings of the 2024 Joint International Conference on Computational
                  Linguistics, Language Resources and Evaluation, {LREC/COLING} 2024,
                  20-25 May, 2024, Torino, Italy},
  pages        = {6426--6430},
  year         = {2024}
}

@inproceedings{Justin2023Closer,
  author       = {Justin Vasselli and
                  Taro Watanabe},
  title        = {A Closer Look at k-Nearest Neighbors Grammatical Error Correction},
  booktitle    = {Proceedings of the 18th Workshop on Innovative Use of {NLP} for Building
                  Educational Applications, BEA@ACL 2023, Toronto, Canada, 13 July 2023},
  pages        = {220--231},
  year         = {2023}
}

@inproceedings{Zhibin2024CRITIC,
  author       = {Zhibin Gou and
                  Zhihong Shao and
                  Yeyun Gong and
                  Yelong Shen and
                  Yujiu Yang and
                  Nan Duan and
                  Weizhu Chen},
  title        = {{CRITIC:} Large Language Models Can Self-Correct with Tool-Interactive
                  Critiquing},
  booktitle    = {The Twelfth International Conference on Learning Representations,
                  {ICLR} 2024, Vienna, Austria, May 7-11, 2024},
  year         = {2024}
}

@article{Liangming2024Automatically,
  author       = {Liangming Pan and
                  Michael Saxon and
                  Wenda Xu and
                  Deepak Nathani and
                  Xinyi Wang and
                  William Yang Wang},
  title        = {Automatically Correcting Large Language Models: \emph{Surveying the
                  Landscape of Diverse Automated Correction Strategies}},
  journal      = {Trans. Assoc. Comput. Linguistics},
  volume       = {12},
  pages        = {484--506},
  year         = {2024}
}

@article{Ryo2024When,
  author       = {Ryo Kamoi and
                  Yusen Zhang and
                  Nan Zhang and
                  Jiawei Han and
                  Rui Zhang},
  title        = {When Can LLMs \emph{Actually} Correct Their Own Mistakes? {A} Critical
                  Survey of Self-Correction of LLMs},
  journal      = {Trans. Assoc. Comput. Linguistics},
  volume       = {12},
  pages        = {1417--1440},
  year         = {2024}
}

@inproceedings{Yaxin2023GrammarGPT,
  author       = {Yaxin Fan and
                  Feng Jiang and
                  Peifeng Li and
                  Haizhou Li},
  title        = {GrammarGPT: Exploring Open-Source LLMs for Native Chinese Grammatical
                  Error Correction with Supervised Fine-Tuning},
  booktitle    = {Natural Language Processing and Chinese Computing - 12th National
                  {CCF} Conference, {NLPCC} 2023, Foshan, China, October 12-15, 2023,
                  Proceedings, Part {III}},
  volume       = {14304},
  pages        = {69--80},
  year         = {2023}
}

@inproceedings{Mengsay2023Exploring,
  author       = {Mengsay Loem and
                  Masahiro Kaneko and
                  Sho Takase and
                  Naoaki Okazaki},
  title        = {Exploring Effectiveness of {GPT-3} in Grammatical Error Correction:
                  {A} Study on Performance and Controllability in Prompt-Based Methods},
  booktitle    = {Proceedings of the 18th Workshop on Innovative Use of {NLP} for Building Educational Applications, BEA@ACL 2023, Toronto, Canada, 13 July 2023},
  pages        = {205--219},
  year         = {2023}
}

@article{Ryo2024Evaluating,
    title = {Evaluating LLMs at Detecting Errors in LLM Responses},
    author = {Ryo Kamoi and Sarkar Snigdha Sarathi Das and Renze Lou and Jihyun Janice Ahn and
      Yilun Zhao and Xiaoxin Lu and Nan Zhang and Yusen Zhang and Ranran Haoran Zhang and
      Sujeeth Reddy Vummanthala and Salika Dave and Shaobo Qin and
      Arman Cohan and Wenpeng Yin and Rui Zhang},
    year = {2024},
    journal = {2024 Conference on Language Modeling},
}

@inproceedings{Masahiro2022Interpretability,
  author       = {Masahiro Kaneko and
                  Sho Takase and
                  Ayana Niwa and
                  Naoaki Okazaki},
  title        = {Interpretability for Language Learners Using Example-Based Grammatical
                  Error Correction},
  booktitle    = {Proceedings of the 60th Annual Meeting of the Association for Computational
                  Linguistics (Volume 1: Long Papers), {ACL} 2022, Dublin, Ireland,
                  May 22-27, 2022},
  pages        = {7176--7187},
  year         = {2022}
}

@inproceedings{ye2025excgec,
  title={EXCGEC: A Benchmark for Edit-Wise Explainable Chinese Grammatical Error Correction},
  author={Ye, Jingheng and Qin, Shang and Li, Yinghui and Cheng, Xuxin and Qin, Libo and Zheng, Hai-Tao and Shen, Ying and Xing, Peng and Xu, Zishan and Cheng, Guo and others},
  booktitle={Proceedings of the AAAI Conference on Artificial Intelligence},
  volume={39},
  number={24},
  pages={25678--25686},
  year={2025}
}

@inproceedings{peng2025encode,
  title={Encode Errors: Representational Retrieval of In-Context Demonstrations for Multilingual Grammatical Error Correction},
  author={Peng, Guangyue and Li, Wei and Luo, Wen and Wang, Houfeng},
  booktitle={Findings of the Association for Computational Linguistics: ACL 2025},
  pages={21166--21180},
  year={2025}
}

@inproceedings{lin2004rouge,
  title={Rouge: A package for automatic evaluation of summaries},
  author={Lin, Chin-Yew},
  booktitle={Text summarization branches out},
  pages={74--81},
  year={2004}
}

@inproceedings{Tianyi2020BERTScore,
  author       = {Tianyi Zhang and
                  Varsha Kishore and
                  Felix Wu and
                  Kilian Q. Weinberger and
                  Yoav Artzi},
  title        = {BERTScore: Evaluating Text Generation with {BERT}},
  booktitle    = {8th International Conference on Learning Representations, {ICLR} 2020,
                  Addis Ababa, Ethiopia, April 26-30, 2020},
  year         = {2020}
}

@inproceedings{Thibault2020BLEURT,
  author       = {Thibault Sellam and
                  Dipanjan Das and
                  Ankur P. Parikh},
  title        = {{BLEURT:} Learning Robust Metrics for Text Generation},
  booktitle    = {Proceedings of the 58th Annual Meeting of the Association for Computational
                  Linguistics, {ACL} 2020, Online, July 5-10, 2020},
  pages        = {7881--7892},
  year         = {2020}
}

@inproceedings{abacha2024medec,
  author       = {Asma Ben Abacha and
                  Wen{-}wai Yim and
                  Yujuan Fu and
                  Zhaoyi Sun and
                  Meliha Yetisgen and
                  Fei Xia and
                  Thomas Lin},
  title        = {{MEDEC:} {A} Benchmark for Medical Error Detection and Correction
                  in Clinical Notes},
  booktitle    = {Findings of the Association for Computational Linguistics, {ACL} 2025,
                  Vienna, Austria, July 27 - August 1, 2025},
  pages        = {22539--22550},
year         = {2025}
}

@inproceedings{abacha2024overview,
  title={Overview of the mediqa-corr 2024 shared task on medical error detection and correction},
  author={Abacha, Asma Ben and Yim, Wen-wai and Fu, Yujuan and Sun, Zhaoyi and Xia, Fei and Yetisgen-Yildiz, Meliha},
  booktitle={Proceedings of the 6th Clinical Natural Language Processing Workshop},
  pages={596--603},
  year={2024}
}

@article{jin2021disease,
  title={What disease does this patient have? a large-scale open domain question answering dataset from medical exams},
  author={Jin, Di and Pan, Eileen and Oufattole, Nassim and Weng, Wei-Hung and Fang, Hanyi and Szolovits, Peter},
  journal={Applied Sciences},
  pages={6421},
  year={2021},
}

@article{daoud2025medarabiq,
  title={Medarabiq: Benchmarking large language models on arabic medical tasks},
  author={Daoud, Mouath Abu and Abouzahir, Chaimae and Kharouf, Leen and Al-Eisawi, Walid and Habash, Nizar and Shamout, Farah E},
  journal={arXiv preprint arXiv:2505.03427},
  year={2025}
}

@inproceedings{wu2024knowlab_aimed,
  title={KnowLab\_AIMed at MEDIQA-CORR 2024: Chain-of-Though (CoT) prompting strategies for medical error detection and correction},
  author={Wu, Zhaolong and Hasan, Abul and Wu, Jinge and Kim, Yunsoo and Cheung, Jason and Zhang, Teng and Wu, Honghan},
  booktitle={proceedings of the 6th clinical natural language processing workshop},
  pages={353--359},
  year={2024}
}

@inproceedings{gundabathula2024promptmind,
  author       = {Satya Kesav Gundabathula and
                  Sriram R. Kolar},
  title        = {PromptMind Team at {MEDIQA-CORR} 2024: Improving Clinical Text Correction
                  with Error Categorization and {LLM} Ensembles},
  booktitle    = {Proceedings of the 6th Clinical Natural Language Processing Workshop,
                  ClinicalNLP@NAACL 2024, Mexico City, Mexico, June 21, 2024},
  pages        = {367--373},
  year         = {2024}
}

@inproceedings{rajwal2024em_mixers,
  title={EM\_Mixers at MEDIQA-CORR 2024: Knowledge-Enhanced Few-Shot In-Context Learning for Medical Error Detection and Correction},
  author={Rajwal, Swati and Agichtein, Eugene and Sarker, Abeed},
  booktitle={Proceedings of the 6th Clinical Natural Language Processing Workshop},
  year={2024}
}

@inproceedings{saeed2024medifact,
  author       = {Nadia Saeed},
  title        = {MediFact at {MEDIQA-CORR} 2024: Why {AI} Needs a Human Touch},
  booktitle    = {Proceedings of the 6th Clinical Natural Language Processing Workshop,
                  ClinicalNLP@NAACL 2024, Mexico City, Mexico, June 21, 2024},
  pages        = {346--352},
  year         = {2024}
}

@inproceedings{corbeil2024iryonlp,
  author       = {Jean{-}Philippe Corbeil},
  editor       = {Tristan Naumann and
                  Asma Ben Abacha and
                  Steven Bethard and
                  Kirk Roberts and
                  Danielle S. Bitterman},
  title        = {IryoNLP at {MEDIQA-CORR} 2024: Tackling the Medical Error Detection
                  {\&} Correction Task on the Shoulders of Medical Agents},
  booktitle    = {Proceedings of the 6th Clinical Natural Language Processing Workshop,
                  ClinicalNLP@NAACL 2024, Mexico City, Mexico, June 21, 2024},
  pages        = {570--580},
  year         = {2024}
}

@inproceedings{valiev2024hse,
  title={HSE NLP team at MEDIQA-CORR 2024 task: In-prompt ensemble with entities and knowledge graph for medical error correction},
  author={Valiev, Airat and Tutubalina, Elena},
  booktitle={Proceedings of the 6th Clinical Natural Language Processing Workshop},
  year={2024}
}

@misc{openai2024gpt4o,
  author = {{OpenAI}},
  title = {GPT-4o Technical Overview},
  year = {2024},
  howpublished = {\url{https://platform.openai.com/docs/models#gpt-4o}},
  note = {Accessed: 2025-08-01}
}

@misc{openai2024gpt4omini,
  author = {{OpenAI}},
  title = {GPT-4o-mini Model Card},
  year = {2024},
  howpublished = {\url{https://research.google/blog/medgemma-our-most-capable-open-models-for\ -health-ai-development}},
  note = {Accessed: 2025-08-01}
}

@misc{google2025medgemma,
  author = {{Google Research}},
  title = {MedGemma: Open Multimodal Models for Medical AI},
  year = {2025},
  howpublished = {\url{https://research.google/blog/medgemma-our-most-capable-open-models-for\ -health-ai-development}},
  note = {Accessed: 2025-08-01}
}

@misc{meta2024llama3,
  author = {{Meta AI}},
  title = {LLaMA 3: Open Foundation and Instruction Models},
  year = {2024},
  howpublished = {\url{https://github.com/meta-llama/llama3}},
  note = {Accessed: 2025-08-01}
}

@misc{qwen2024,
  author = {{Alibaba DAMO Academy}},
  title = {Qwen2.5-7B-Instruct-1M},
  year = {2024},
  howpublished = {\url{https://huggingface.co/Qwen/Qwen2.5-7B-Instruct}},
  note = {Accessed: 2025-08-01}
}

@misc{deepseekr12024,
  author = {{DeepSeek AI}},
  title = {DeepSeek-R1},
  year = {2024},
  howpublished = {\url{https://huggingface.co/deepseek-ai/deepseek-llm-7b-base}},
  note = {Accessed: 2025-08-01}
}

@misc{deepseekv32024,
  author = {{DeepSeek AI}},
  title = {DeepSeek-V3},
  year = {2024},
  howpublished = {\url{https://huggingface.co/deepseek-ai/deepseek-vl-7b}},
  note = {Accessed: 2025-08-01}
}

@misc{doubao2024,
  author = {{Doubao Team, Volcano Engine}},
  title = {Doubao-1.5-Thinking-Pro},
  year = {2024},
  howpublished = {\url{https://console.volcengine.com/ark/region:ark+cn-beijing/model/detail?Id=doubao-1-5-thinking-pro}},
  note = {Accessed: 2025-08-01}
}

@inproceedings{bari2025allam,
  author = {Bari, M. Saiful and Alnumay, Yazeed and Alzahrani, Norah A. and others},
  title = {ALLaM: Large Language Models for Arabic and English},
  booktitle = {ICLR},
  year = {2025},
  howpublished = {\url{https://arxiv.org/abs/2407.15390}}
}

@misc{huatuogpt2024,
      title={HuatuoGPT-o1, Towards Medical Complex Reasoning with LLMs}, 
      author={Junying Chen and Zhenyang Cai and Ke Ji and Xidong Wang and Wanlong Liu and Rongsheng Wang and Jianye Hou and Benyou Wang},
      year={2024},
      eprint={2412.18925},
      archivePrefix={arXiv},
      primaryClass={cs.CL},
      url={https://arxiv.org/abs/2412.18925}, 
}

@misc{medgemma2025blog,
  author = {{Google Research}},
  title = {MedGemma: our most capable open models for health AI},
  year = {2025},
  howpublished = {\url{https://research.google/blog/medgemma-our-most-capable-open-models-for\ -health-ai-development}},
  note = {Accessed: 2025-08-01}
}

\appendix
 \setcounter{table}{0}
\renewcommand{\thetable}{S\arabic{table}}

 \setcounter{figure}{0}
\renewcommand{\thefigure}{S\arabic{figure}}

\section{Data Statistics}

\subsection{Language-wise data split and statistics}
Table \ref{tab: statistics} presents the official data splits and basic statistics of the  dataset, a multilingual clinical benchmark covering English, Chinese, and Arabic. For each language, we report the number of instances in the training, validation, and test sets, along with the total number of samples. The English dataset contains 1,024 instances, the Chinese dataset includes 1,000, and the Arabic dataset consists of 482. We also provide average, maximum, and minimum input lengths, as well as the number of samples with and without factual errors in each split.

\begin{table}[ht] 
\centering
\caption{Summary Statistics of the  Dataset.}
\resizebox{0.48\textwidth}{!}{%
\begin{tabular}{llcccc}
\toprule
\textbf{Language} & \textbf{Metric} & \textbf{Train} & \textbf{Validation} & \textbf{Test} & \textbf{Total} \\
\midrule
\multirow{6}{*}{\textbf{English}} 
 & Num. & 708 & 108 & 208 & 1024 \\
 & Avg len & 755.7 & 604.3 & 872.1 & 763.9 \\
 & Max len & 1594 & 1250 & 1396 & 1594 \\
 & Min len & 220 & 101 & 63 & 63 \\
 & w errors & 354   & 54   & 104   & 512   \\
 & w/o errors & 354   & 54   & 104   & 512   \\
\midrule
\multirow{6}{*}{\textbf{Chinese}} 
 & Num. & 700 & 100 & 200 & 1000 \\
 & Avg len & 97.4 & 96.3 & 98 & 97.3 \\
 & Max len & 269 & 262 & 191 & 269 \\
 & Min len & 32 & 41 & 40 & 32 \\
 & w errors & 350   & 50   & 100   & 500   \\
 & w/o errors & 350   & 50   & 100   & 500   \\
\midrule
\multirow{6}{*}{\textbf{Arabic}} 
 & Num. & 334 & 51 & 97 & 482 \\
 & Avg len & 156 & 168.2 & 165.6 & 159.2 \\
 & Max len & 457 & 429 & 501 & 501 \\
 & Min len & 31 & 48 & 37 & 31 \\
 & w errors & 179  & 29  & 53  & 261  \\
 & w/o errors & 155  & 22  & 44  & 221  \\
\bottomrule
\end{tabular}%
}
\label{tab: statistics}
\end{table}

\subsection{Distribution of medical error types}
\begin{figure*}[t]
    \centering
    \includegraphics[width=1\textwidth]{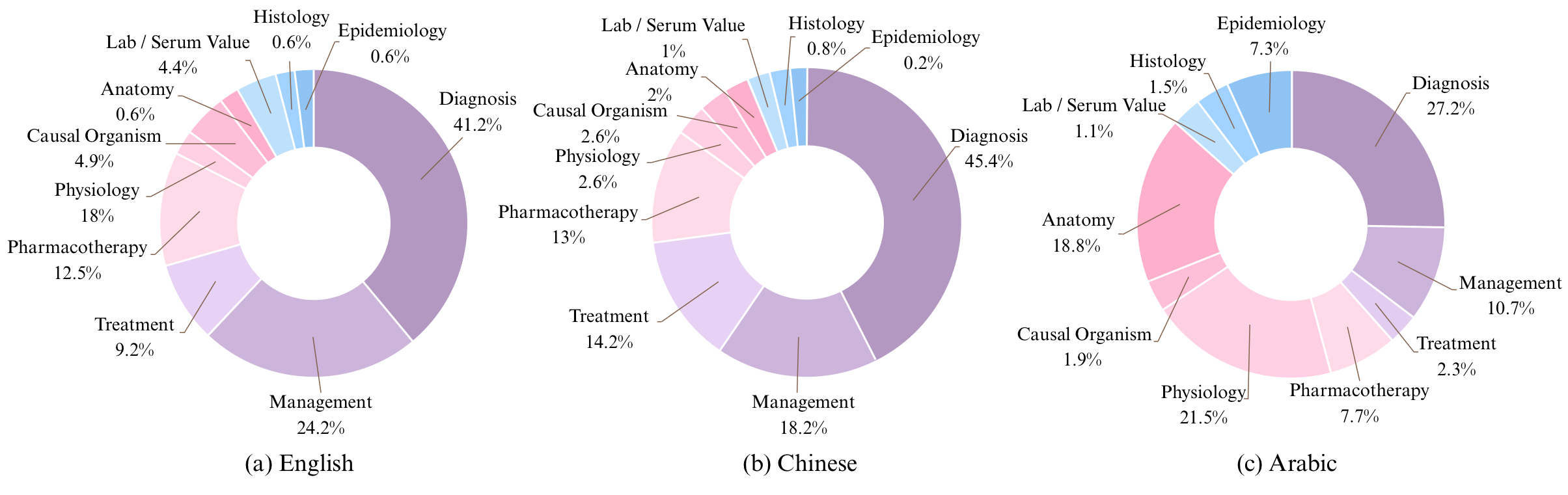}
    \vspace{-2mm}
    \caption{Distribution of Error Types by Language}
    \label{fig: Error_Distribution}
\end{figure*}

Figure \ref{fig: Error_Distribution} illustrates the distribution of ten common medical error types in the  dataset across three languages: English, Chinese, and Arabic. Each donut chart represents the relative proportions of error categories, including Diagnosis, Management, Treatment, Pharmacotherapy, Physiology, Causal Organism, Anatomy, Lab/Serum Value, Histology, and Epidemiology. In the English subset (a), Diagnosis (41.2\%) and Management (24.2\%) are the most prevalent error types, followed by Physiology (18.0\%) and Pharmacotherapy (12.5\%). The Chinese subset (b) shows a stronger concentration in Diagnosis errors (45.4\%), with moderate representations of Mangement (18.2\%), Treatment (14.2\%), Pharmacotherapy (13.0\%), and notably fewer errors related to Physiology and Anatomy (2.6\% each). In contrast, the Arabic subset (c) presents a more balanced distribution, where Diagnosis (27.2\%) remains the largest category, but Physiology (21.5\%) and Anatomy (18.8\%) are more prominent, while categories such as Treatment (2.3\%) and Causal Organism (1.9\%) are less frequent. These distributions highlight linguistic and potentially systemic differences in error typologies across multilingual medical corpora.

\section{Benchmark Tasks}

The proposed MedErrBench dataset supports three core tasks for benchmarking clinical error understanding:

\begin{itemize}

\item Error Detection: Determine whether a given clinical note contains an error. This is formulated as a binary classification task distinguishing between error-free and erroneous notes.

\item Error Localization: Identify which specific sentence within the clinical note contains the error. This task focuses on sentence-level localization rather than token-level span extraction, aligning with how clinicians typically review clinical documentation.

\item Error Correction: Generate a revised version of the clinical note with the error corrected. This task requires contextual understanding and clinical knowledge to produce plausible and medically valid corrections.

\end{itemize}

Beyond the three primary tasks supported by MedErrBench, the dataset includes annotations of difficulty level, reasoning type, and important clinical terms, which enable flexible task customization. For instance, MedErrBench can be used for error classification by assigning each clinical note to a predefined error category curated by expert clinicians. In addition, annotated key clinical terms support token-level localization, facilitating the development of alternative task formulations.

\section{Experimental Settings}

During dataset construction, LLMs used for the English, Chinese, and Arabic datasets were Gemini 2.0 Flash, DeepSeek-V3-0324, and Gemini-1.5-pro, respectively.
In regard to the models, 
Qwen2.5-7B-Instruct was used and all the hyperparameters were set to the default values\footnote{\url{https://www.alibabacloud.com/help/en/model-studio/use-qwen-by-calling-api}}. ALLAM-7B was run with hyperparameters: max\_new\_tokens=256, top\_k=50, temperature=0.7, top\_p=0.9, and do\_sample=True. JAIS-adapted-13b-chat was run wiht max\_new\_tokens=256, do\_sample=True,  temperature=0.7, top\_k = 50 and top\_p=0.95. Doubao-1.5-Thinking-Pro was run with max\_tokens=512, with all other parameters (including temperature and top\_p) set to default values\footnote{\url{https://www.volcengine.com/docs/82379/1494384}}. DeepSeek-V3 and DeepSeek-R-1 were run with hyperparameters: max\_tokens=512, while all other parameters were set to their default values\footnote{\url{api-docs.deepseek.com/api/create-chat-completion}}. Meta-Llama-3.1-8B-Instruct was accessed via the Novita InferenceClient with hyperparameters: max\_tokens=256, temperature=0.2, do\_sample=False. HuatuoGPT-o1-8B was run locally with PyTorch with hyperparameters: max\_new\_tokens=1024, temperature=0.2, do\_sample=False, and pad\_token\_id set to the tokenizer’s eos\_token\_id. GPT-4o, GPT-4o-mini, Gemini 2.0 Flash, Gemini 2.5 Flash Lite, and MedGemma-4B, MedGemma-27B are all used with default parameters.
We use BLEURT with BLEURT-20 model\footnote{\url{https://github.com/google-research/bleurt}} and BERTScore with the deberta-xlarge-mnli checkpoint\footnote{\url{https://huggingface.co/microsoft/deberta-xlarge-mnli}}.
For Chinese data, ROUGE evaluation is performed using the ROUGE-chinese package\footnote{\url{https://pypi.org/project/rouge-chinese/}}, which provides tokenization and evaluation methods specifically designed for Chinese text.

\section{Evaluation Metrics}
ROUGE-1 \cite{lin2004rouge} measures unigram overlap between the generated and reference texts, while ROUGE-2 extends this to bigram overlap. ROUGE-L \cite{lin2004rouge} captures structural similarity based on the longest common subsequence and ROUGE-SU \cite{lin2004rouge} incorporates both unigrams and skip-bigrams with a maximum skip distance, offering a balance between flexibility and structure. BERTScore \cite{Tianyi2020BERTScore} evaluates semantic similarity using contextual embeddings from a pre-trained BERT model. BLEURT \cite{Thibault2020BLEURT} combines pre-trained language models with human-annotated data to assess the fluency and adequacy of the generated text.

\section{Detailed Overall Performance Across Languages}
We report the complete results on all evaluation metrics, including detection accuracy, localization accuracy, ROUGE-1, ROUGE-2, ROUGE-L, BLEU, BERTScore, and BLEURT, evaluated on MedErrBench-EN (Table \ref{tab:full-clinmed-english-results}), MedErrBench-CN (Table \ref{tab:full-clinmed-chinese-results}), and MedErrBench-Ara (Table \ref{tab:full-clinmed-arabic-results}).

\renewcommand{\arraystretch}{0.85}
\begin{table*}[t]
\centering
\caption{Results on  MedErrBench-EN.}
\label{tab:full-clinmed-english-results}
\resizebox{0.9\textwidth}{!}{%
\begin{tabular}{lcccccccc}
\toprule
\textbf{Models} & \multicolumn{1}{c}{\textbf{Detection}} & \multicolumn{1}{c}{\textbf{Localization}} & \multicolumn{6}{c}{\textbf{Error Correction}} \\
\textbf{} & \textbf{Accuracy} & \textbf{Accuracy} & \textbf{ROUGE-1} & \textbf{ROUGE-2} & \textbf{ROUGE-L} & \textbf{BLEU} & \textbf{BERTScore} & \textbf{BLEURT} \\
\midrule
\rowcolor{gray!15} \multicolumn{9}{c}{\textit{General-purpose LLMs}} \\
\midrule
gpt-4o                & 0.596 & 0.346 & 0.415 & 0.365 & 0.403 & 0.329 & 0.428 & 0.407 \\
gpt-4o-mini  & \underline{0.664} & 0.524 & 0.487 & 0.441 & 0.477 & 0.409 & 0.498 & 0.472          \\
Gemini 2.5 Flash Lite & 0.567 & 0.264 & 0.349 & 0.307 & 0.340 & 0.279 & 0.362 & 0.346 \\
Gemini 2.0 Flash   & 0.514 & 0.168 & 0.281 & 0.231 & 0.268 & 0.186 & 0.294 & 0.288    \\
Llama3-8b             & 0.519 & 0.361 & 0.266 & 0.225 & 0.261 & 0.219 & 0.261 & 0.283 \\
Llama-3.3-70B-Instruct &0.582	&0.255	&0.369	&0.328	&0.356	&0.292	&0.369 &0.385 \\
\midrule
\rowcolor{gray!15} \multicolumn{9}{c}{\textit{Language-specialized LLMs}} \\
\midrule
Qwen2.5-7B-Instruct        & 0.563 & 0.490 & 0.372 & 0.334 & 0.345 & 0.381 & 0.450 & 0.371 \\
Deepseek-R1                   & 0.582 & 0.577 & 0.700 & 0.612 & 0.682 & 0.551 & 0.716 & 0.681 \\
Deepseek-V3                   & 0.587 & \underline{0.582} & \underline{0.703} & \underline{0.626} & \underline{0.687} & \underline{0.569} & \underline{0.732} & \underline{0.693} \\
Doubao-1.5      & \textbf{0.779} & \textbf{0.774} & \textbf{0.766} &\textbf{ 0.707} & \textbf{0.752} &\textbf{ 0.662} & \textbf{0.783 }& \textbf{0.773}\\
ALLAM-7B                        & 0.029 & 0.014 & 0.015 & 0.014 & 0.015 & 0.286 & 0.020 & 0.014 \\
\midrule
\rowcolor{gray!15} \multicolumn{9}{c}{\textit{Medical-domain LLMs}} \\
\midrule
MedGemma-4b       & 0.505 & 0.438 & 0.511 & 0.499 & 0.508 & 0.489 & 0.518 & 0.513 \\
MedGemma-27b      & 0.543 & 0.245 & 0.377 & 0.337 & 0.369 & 0.305 & 0.390 & 0.349 \\
HuatuoGPT-o1-7b   & 0.574 & 0.530 & 0.486 & 0.466 & .485 & 0.446 & 0.475 & 0.475 \\
\bottomrule

\end{tabular}%
}
\end{table*}

\renewcommand{\arraystretch}{0.85}
\begin{table*}[ht]
\centering
\caption{Results on MedErrBench-CN.}
\label{tab:full-clinmed-chinese-results}
\resizebox{0.9\textwidth}{!}{%
\begin{tabular}{lcccccccc}
\toprule
\textbf{Models} & \multicolumn{1}{c}{\textbf{Detection}} & \multicolumn{1}{c}{\textbf{Localization}} & \multicolumn{6}{c}{\textbf{Error Correction}} \\
\textbf{} & \textbf{Accuracy} & \textbf{Accuracy} & \textbf{ROUGE-1} & \textbf{ROUGE-2} & \textbf{ROUGE-L} & \textbf{BLEU} & \textbf{BERTScore} & \textbf{BLEURT} \\
\midrule
\rowcolor{gray!15} \multicolumn{9}{c}{\textit{General-purpose LLMs}} \\
\midrule
gpt-4o                     & 0.630 & 0.205 & 0.265 & 0.169 & 0.247 & 0.148 & 0.365 & 0.266 \\
gpt-4o-mini               & 0.505 & 0.115 & 0.244 & 0.145 & 0.223 & 0.130 & 0.390 & 0.257 \\
Gemini 2.5 Flash Lite     & 0.600 & 0.375 & 0.448 & 0.401 & 0.439 & 0.310 & 0.533 & 0.455 \\
Gemini 2.0 Flash          & 0.705 & 0.455 & 0.569 & 0.510 & 0.557 & 0.405 & 0.659 & 0.577 \\
Llama3-8b                 & 0.500 & 0.320 & 0.416 & 0.251 & 0.377 & 0.168 & 0.532 & 0.483 \\
Llama-3.3-70B-Instruct	&0.675	&0.380	&0.506	&0.459	&0.500	&0.438	&0.606	&0.509 \\
\midrule
\rowcolor{gray!15} \multicolumn{9}{c}{\textit{Language-specialized LLMs}} \\
\midrule
Qwen2.5-7B-Instruct    & 0.625 & 0.570 & 0.493 & 0.429 & 0.492 & 0.331 & 0.576 & 0.462 \\
Deepseek-R1    & \underline{0.735} & \underline{0.705} & \underline{0.802} & 0.708 & \underline{0.799} & 0.491 & \underline{0.851} & \underline{0.781} \\
Deepseek-V3    & 0.650 & 0.640 & \textbf{0.833} & \textbf{0.741} & \textbf{0.830} & \textbf{0.540} & \textbf{0.873} & \textbf{0.806} \\
Doubao-1.5   & \textbf{0.750} & \textbf{0.735 }& 0.788 & \underline{0.709} & 0.784 & \underline{0.508} & 0.835 & 0.777 \\
ALLAM-7B                    & 0.395 & 0.340 & 0.284 & 0.260 & 0.283 & 0.080 & 0.360 & 0.286 \\
\midrule
\rowcolor{gray!15} \multicolumn{9}{c}{\textit{Medical-domain LLMs}} \\
\midrule
MedGemma-4b               & 0.525 & 0.500 & 0.549 & 0.528 & 0.545 & 0.500 & 0.581 & 0.547 \\
MedGemma-27b              & 0.605 & 0.285 & 0.441 & 0.386 & 0.430 & 0.270 & 0.537 & 0.438 \\
HuatuoGPT-o1-7b           & 0.525 & 0.275 & 0.167 & 0.056 & 0.167 & 0.169 & 0.545 & 0.530 \\
\bottomrule
\end{tabular}%
}
\end{table*}

\renewcommand{\arraystretch}{0.85}
\begin{table*}[ht]
\centering
\caption{Results on MedErrBench-Ara. }
\label{tab:full-clinmed-arabic-results}
\resizebox{0.9\textwidth}{!}{%
\begin{tabular}{lcccccccc}
\toprule
\textbf{Models} & \multicolumn{1}{c}{\textbf{Detection}} & \multicolumn{1}{c}{\textbf{Localization}} & \multicolumn{6}{c}{\textbf{Error Correction}} \\
\textbf{} & \textbf{Accuracy} & \textbf{Accuracy} & \textbf{ROUGE-1} & \textbf{ROUGE-2} & \textbf{ROUGE-L} & \textbf{BLEU} & \textbf{BERTScore} & \textbf{BLEURT} \\
\midrule
\rowcolor{gray!15} \multicolumn{9}{c}{\textit{General-purpose LLMs}} \\
\midrule
gpt-4o                & \underline{0.680} & 0.320 & 0.399 & 0.357 & 0.393 & 0.321 & 0.592 & 0.414 \\
gpt-4o-mini           & 0.577 & 0.175 & 0.260 & 0.212 & 0.250 & 0.164 & 0.469 & 0.292 \\
Gemini 2.5 Flash Lite & 0.495 & 0.268 & 0.303 & 0.280 & 0.299 & 0.253 & 0.432 & 0.318 \\
Gemini 2.0 Flash      & 0.598 & 0.299 & 0.315 & 0.281 & 0.308 & 0.251 & 0.503 & 0.332 \\
Llama3-8b             & 0.371 & 0.309 & 0.311 & 0.304 & 0.310 & 0.296 & 0.324 & 0.313 \\
Llama-3.3-70B-Instruct	&0.557	&0.381	&0.412	&0.385	&0.406	&0.364	&0.454	&0.405 \\
\midrule
\rowcolor{gray!15} \multicolumn{9}{c}{\textit{Language-specialized LLMs}} \\
\midrule
Qwen2.5-7B-Instruct            & 0.536 & 0.381 & 0.329 & 0.298 & 0.327 & 0.348 & 0.473 & 0.353 \\
Deepseek-R1            & \textbf{0.711} & \textbf{0.505} & 0.568 & 0.500 & 0.564 & \underline{0.457} & \underline{0.756} & \underline{0.610} \\
Deepseek-V3            & 0.608 & \textbf{0.505} & \textbf{0.677} & \textbf{0.628} & \textbf{0.675} & \textbf{0.592} & \textbf{0.814} & \textbf{0.699} \\
Doubao-1.5             & 0.670 & \textbf{0.505} & \underline{0.582} & \underline{0.510} & \underline{0.574} & 0.452 & 0.736 & 0.583 \\
ALLAM-7B                 & 0.072 & 0.021 & 0.045 & 0.044 & 0.045 & 0.219 & 0.049 & 0.046 \\
\midrule
\rowcolor{gray!15} \multicolumn{9}{c}{\textit{Medical-domain LLMs}} \\
\midrule
MedGemma-4b       & 0.454 & 0.433 & 0.438 & 0.436 & 0.438 & 0.436 & 0.450 & 0.439 \\
MedGemma-27b      & 0.552 & 0.240 & 0.266 & 0.220 & 0.257 & 0.185 & 0.456 & 0.286 \\
HuatuoGPT-o1-7b   & 0.371 & 0.397 & 0.351	&0.260	&0.329	&0.158	&0.450	&0.420 \\
\bottomrule
\end{tabular}%
}
\end{table*}

\section{Detailed Experimental Results on the Impact of Providing Example Difficulty Levels in Few-Shot Learning Settings}

Table~\ref{tab:full-error-type-analysis} presents the complete results across all evaluation metrics, including detection accuracy, localization accuracy, ROUGE-1/2/L, BLEU, BERTScore, and BLEURT, for analyzing the impact of providing example difficulty levels in few-shot learning settings.

\renewcommand{\arraystretch}{0.85}
\begin{table*}[t!]
\centering
\caption{Performance comparison of models under different error type Conditions. ``ET"  and ``DEF" indicate error types and definitions, respectively.}
\label{tab:full-error-type-analysis}
\resizebox{0.9\textwidth}{!}{%
\begin{tabular}{lcccccccc}
\toprule
\textbf{} & \multicolumn{1}{c}{\textbf{Detection}} & \multicolumn{1}{c}{\textbf{Localization}} & \multicolumn{6}{c}{\textbf{Error Correction}} \\
\textbf{} & \textbf{Accuracy} & \textbf{Accuracy} & \textbf{ROUGE-1} & \textbf{ROUGE-2} & \textbf{ROUGE-L} & \textbf{BLEU} & \textbf{BERTScore} & \textbf{BLEURT} \\
\midrule
\rowcolor{gray!15} \multicolumn{9}{c}{\textit{Deepseek-V3 (Zero-shot)}} \\
\midrule
\textit{w/o ET \& DEF}
      & 0.690 & 0.645 & 0.660 & 0.490 & 0.659 & 0.192 & 0.735 & 0.599 \\
\textit{w/o DEF}     & 0.625 & 0.610 & 0.731 & 0.564 & 0.726 & 0.227 & 0.795 & 0.685 \\
\textit{  w ET \& DEF}   & 0.650 & 0.640 & 0.730 & 0.574 & 0.724 & 0.228 & 0.794 & 0.684 \\
\midrule
\rowcolor{gray!15} \multicolumn{9}{c}{\textit{Deepseek-V3 (Few-shot)}} \\
\midrule
\textit{w/o ET \& DEF}      & 0.720 & 0.690 & 0.695 & 0.535 & 0.695 & 0.249 & 0.767 & 0.639 \\
\textit{w/o DEF}     & 0.710 & 0.705 & 0.736 & 0.561 & 0.728 & 0.215 & 0.796 & 0.684 \\
\textit{  w ET \& DEF}   & 0.715 & 0.715 & 0.763 & 0.596 & 0.758 & 0.234 & 0.821 & 0.705 \\
\midrule
\rowcolor{gray!15} \multicolumn{9}{c}{\textit{Doubao-1.5-thinking-pro (Zero-shot)}} \\
\midrule
\textit{w/o ET \& DEF}      & 0.695 & 0.640 & 0.637 & 0.503 & 0.632 & 0.166 & 0.727 & 0.607 \\
\textit{w/o DEF}     & 0.730 & 0.710 & 0.673 & 0.521 & 0.665 & 0.164 & 0.751 & 0.646 \\
\textit{  w ET \& DEF}   & 0.750 & 0.725 & 0.669 & 0.531 & 0.660 & 0.171 & 0.728 & 0.636 \\
\midrule
\rowcolor{gray!15} \multicolumn{9}{c}{\textit{Doubao-1.5-thinking-pro (Few-shot)}} \\
\midrule
\textit{w/o ET \& DEF}     & 0.735 & 0.695 & 0.707 & 0.553 & 0.695 & 0.248 & 0.765 & 0.651 \\
\textit{w/o DEF}     & 0.765 & 0.750 & 0.729 & 0.568 & 0.716 & 0.260 & 0.777 & 0.671 \\
\textit{  w ET \& DEF}   & 0.775 & 0.765 & 0.699 & 0.535 & 0.687 & 0.209 & 0.753 & 0.665 \\
\bottomrule
\end{tabular}%
}
\end{table*}

\section{Prompts of Data Construction}

\begin{figure*}[t]
    \centering
    \includegraphics[width=1\textwidth]{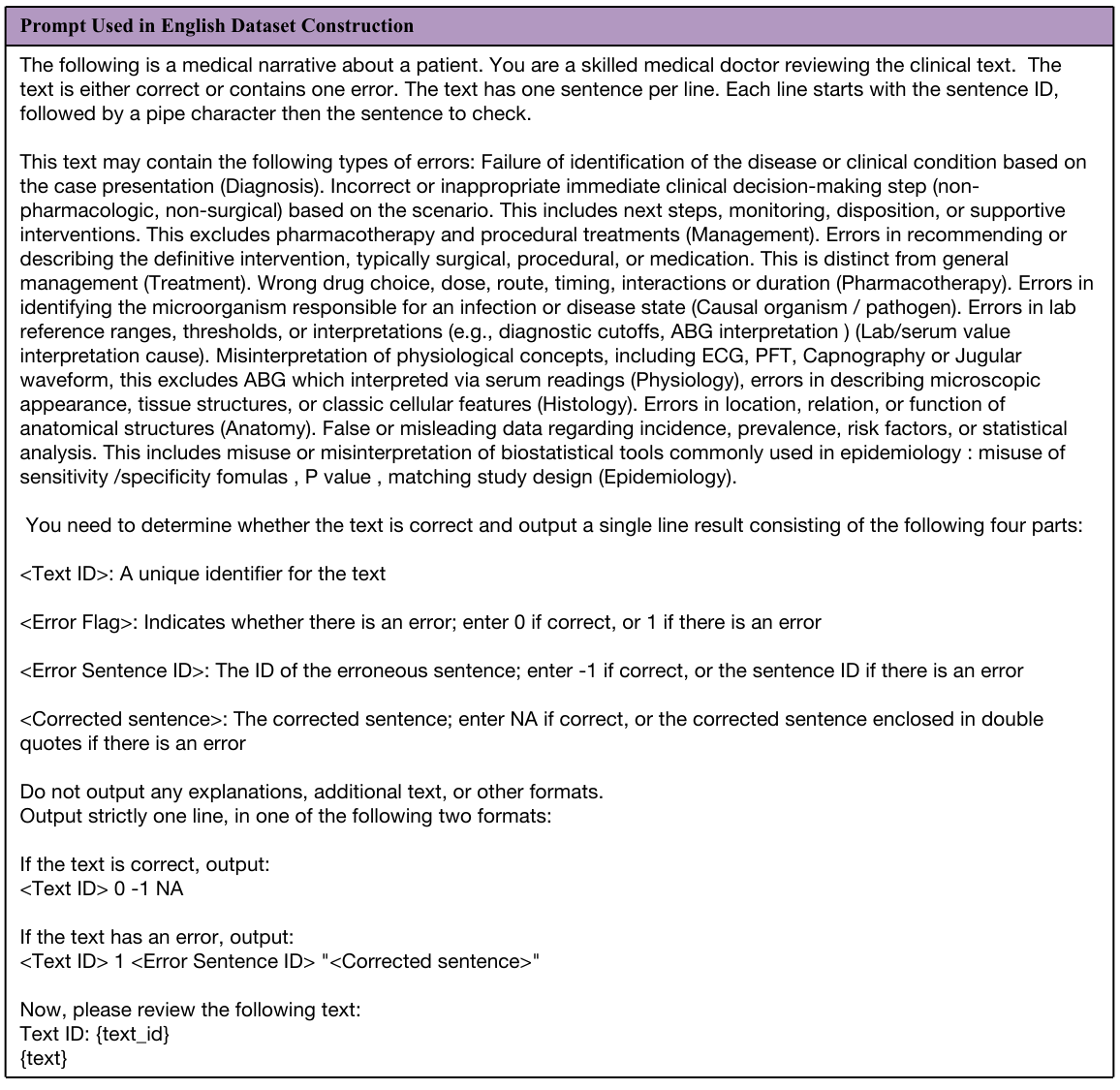}
    \vspace{-2mm}
    \caption{Prompt Used in MedErrBench-En Construction.}
    \label{fig: Prompt_EN}
\end{figure*}

\begin{figure*}[t]
    \centering
    \includegraphics[width=1\textwidth]{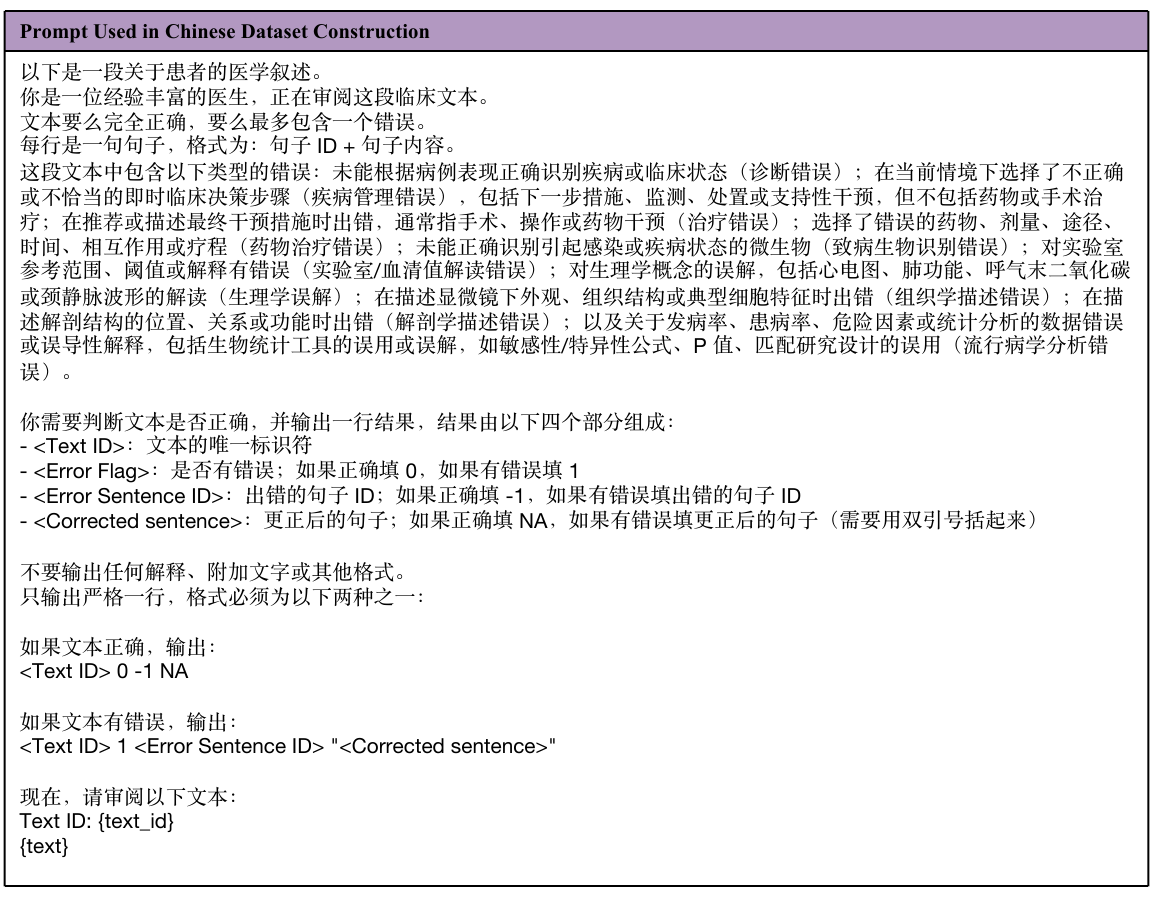}
    \vspace{-2mm}
    \caption{Prompt Used in MedErrBench-CN Construction.}
    \label{fig: Prompt_CN}
\end{figure*}

\begin{figure*}[t]
    \centering
    \includegraphics[width=1\textwidth]{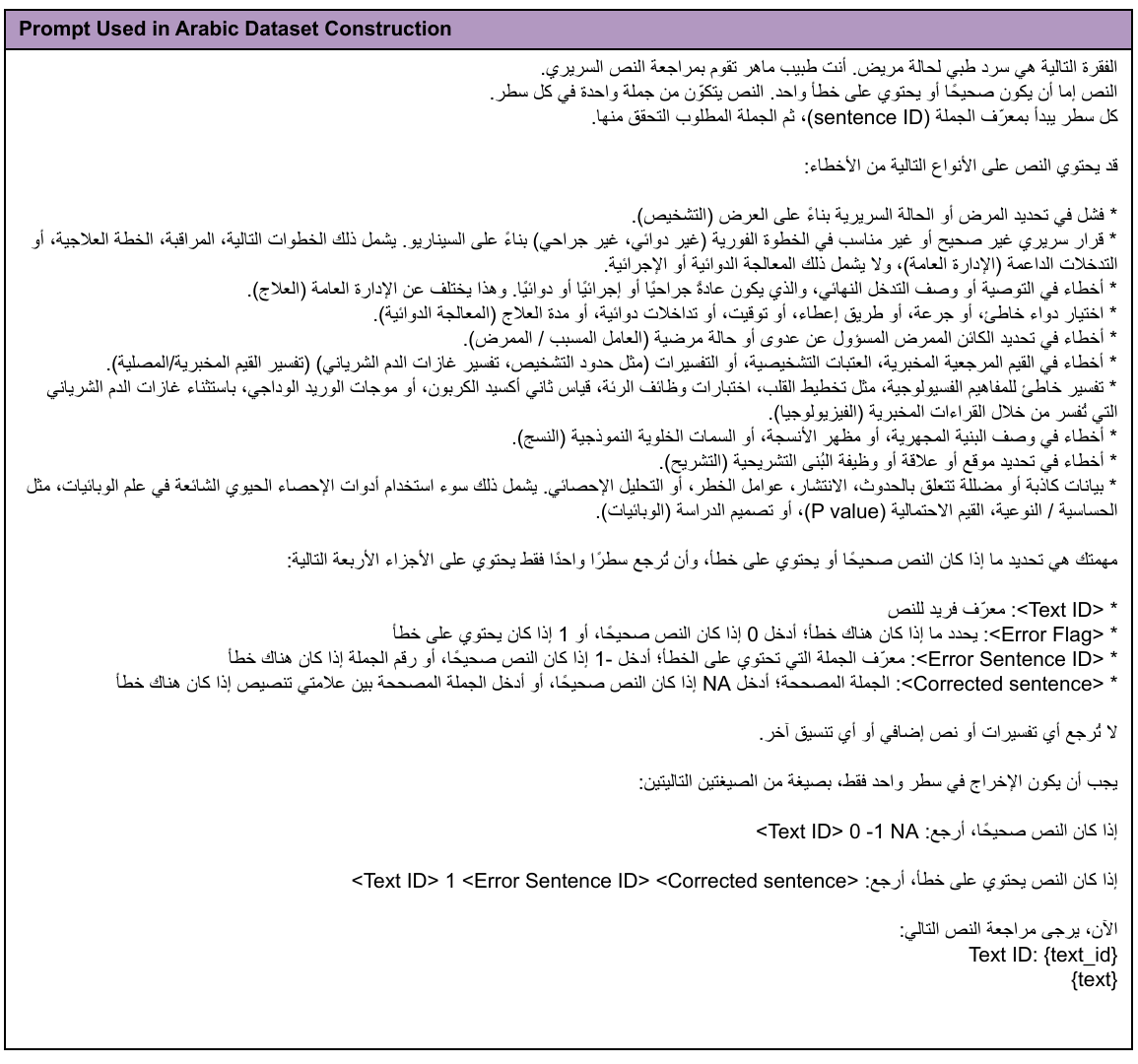}
    \vspace{-2mm}
    \caption{Prompt Used in MedErrBench-Ara Construction.}
    \label{fig: Prompt_Ara}
\end{figure*}

We utilized large language models to help convert medical examination materials into comprehensive clinical stories across English, Chinese, and Arabic. The language-specific prompts used for generation are illustrated in Figure \ref{fig: Prompt_EN}, Figure \ref{fig: Prompt_CN} and \ref{fig: Prompt_Ara}.

\section{Examples for Error Insertion in MedErrBench Dataset}

\begin{figure*}[t]
    \centering
    \includegraphics[width=1\textwidth]{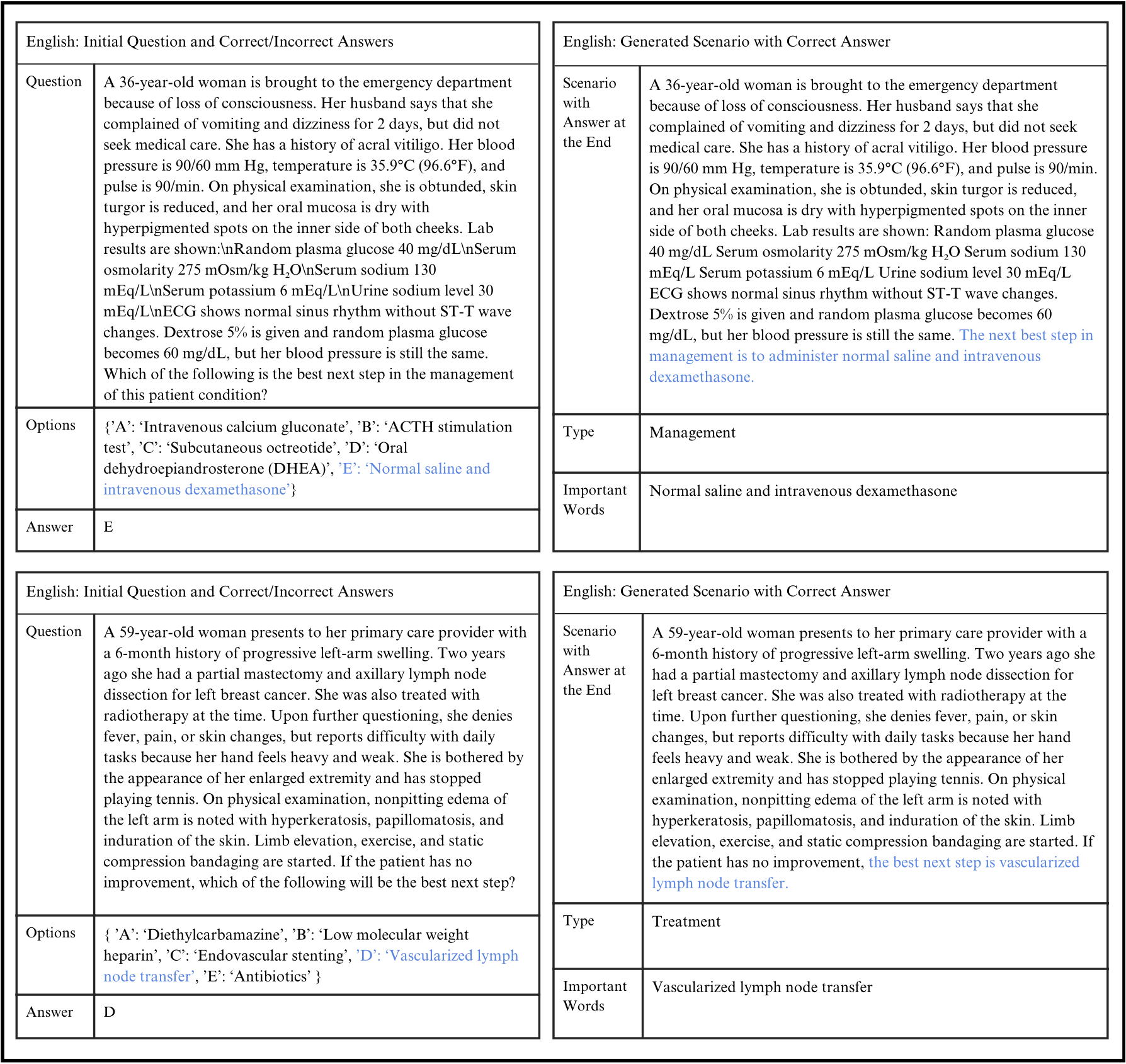}
    \vspace{-2mm}
    \caption{Examples for Error Insertion in MedErrBench-EN.}
    \label{fig:Example_EN}
\end{figure*}

\begin{figure*}[t]
    \centering
    \includegraphics[width=1\textwidth]{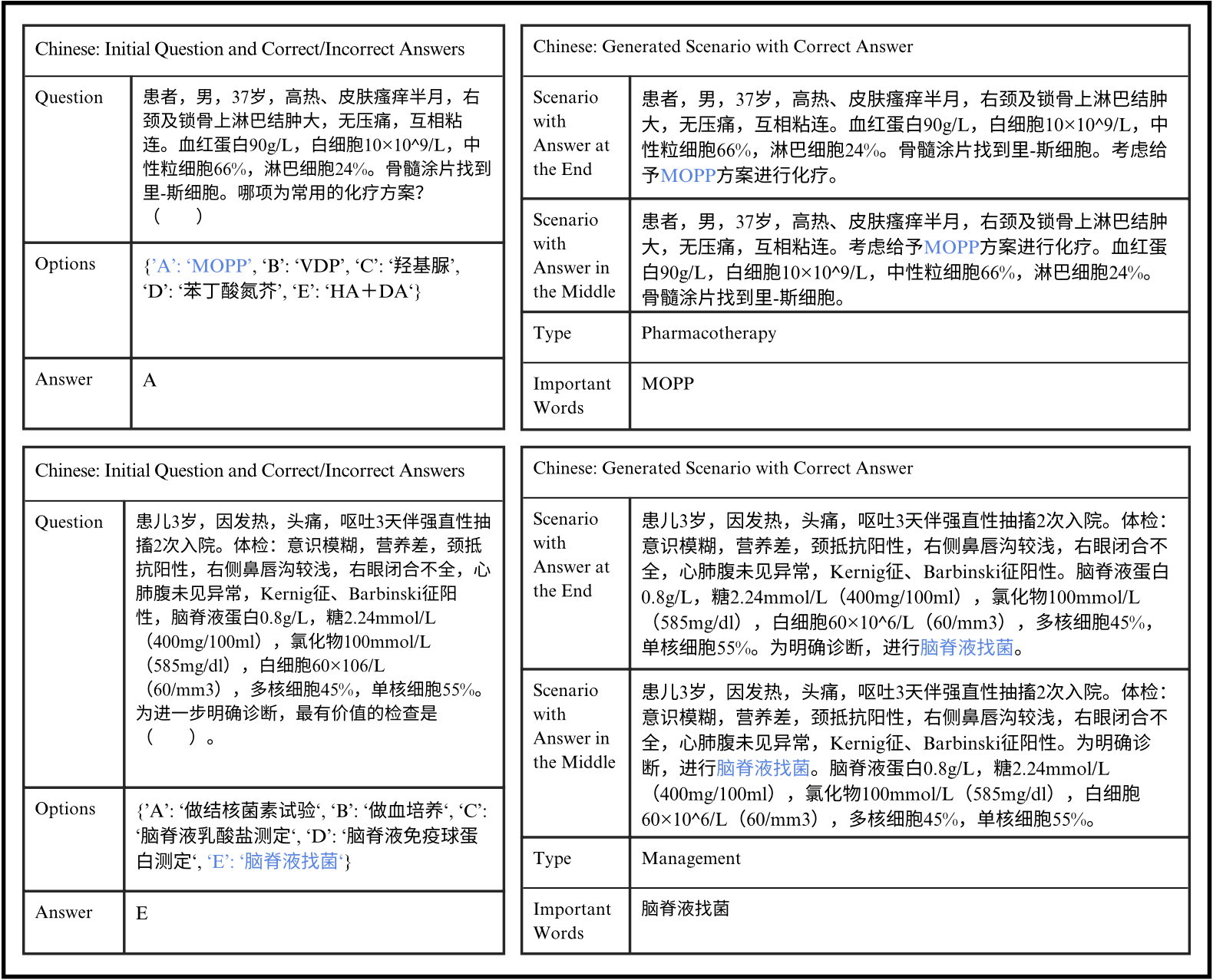}
    \vspace{-2mm}
    \caption{Examples for Error Insertion in MedErrBench-CN.}
    \label{fig:Example_CN}
\end{figure*}

\begin{figure*}[t]
    \centering
    \includegraphics[width=0.8\textwidth]{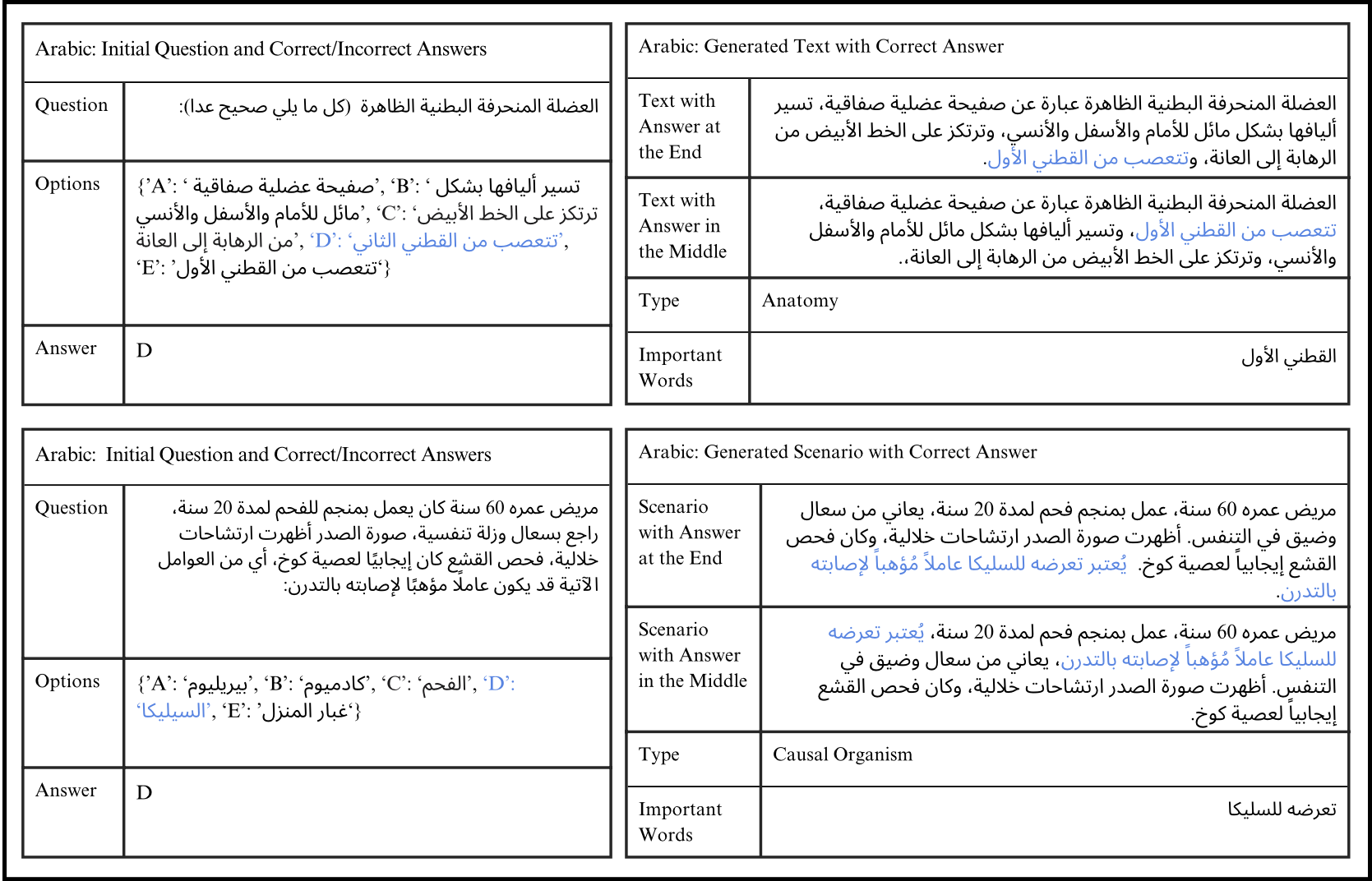}
    \vspace{-2mm}
    \caption{Examples for Error Insertion in MedErrBench-Ara.}
    \label{fig:Example_Ara}
\end{figure*}

Figure \ref{fig:Example_EN}, \ref{fig:Example_CN} and \ref{fig:Example_Ara} are examples for error insertion in the proposed MedErrBench dataset.

\section {Illustrative Examples of the Ten Error Types Defined in the MedErrBench Dataset}

\begin{figure*}[t]
    \centering
    \includegraphics[width=1\textwidth,height=0.8\textheight,keepaspectratio]{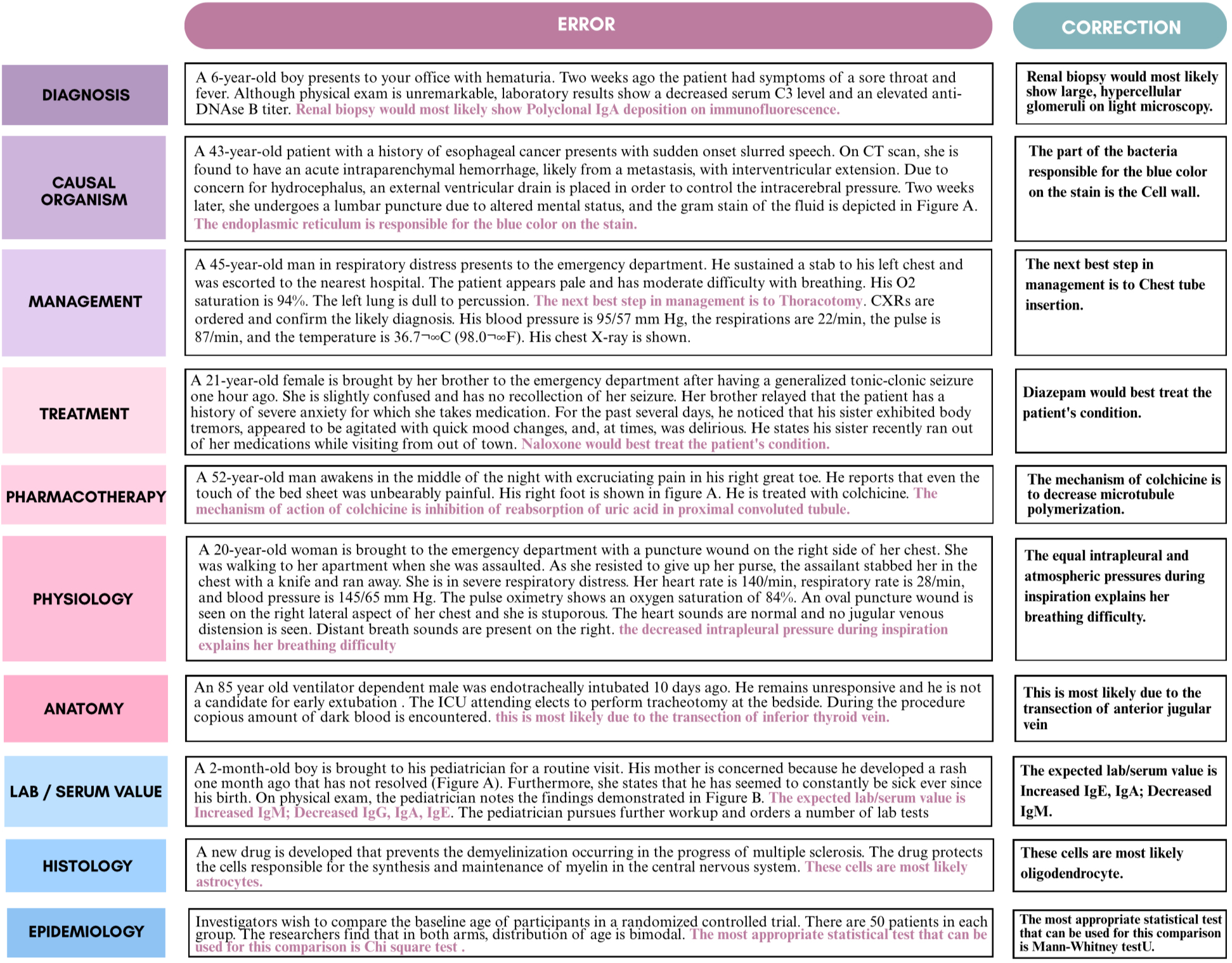}
    \vspace{-2mm}
    \caption{Illustrative Examples of the Ten Error Types Defined in the MedErrBench-EN.}
    \label{fig:Error_type_En}
\end{figure*}

\begin{figure*}[t]
    \centering
    \includegraphics[width=0.8\textwidth,height=0.8\textheight,keepaspectratio]{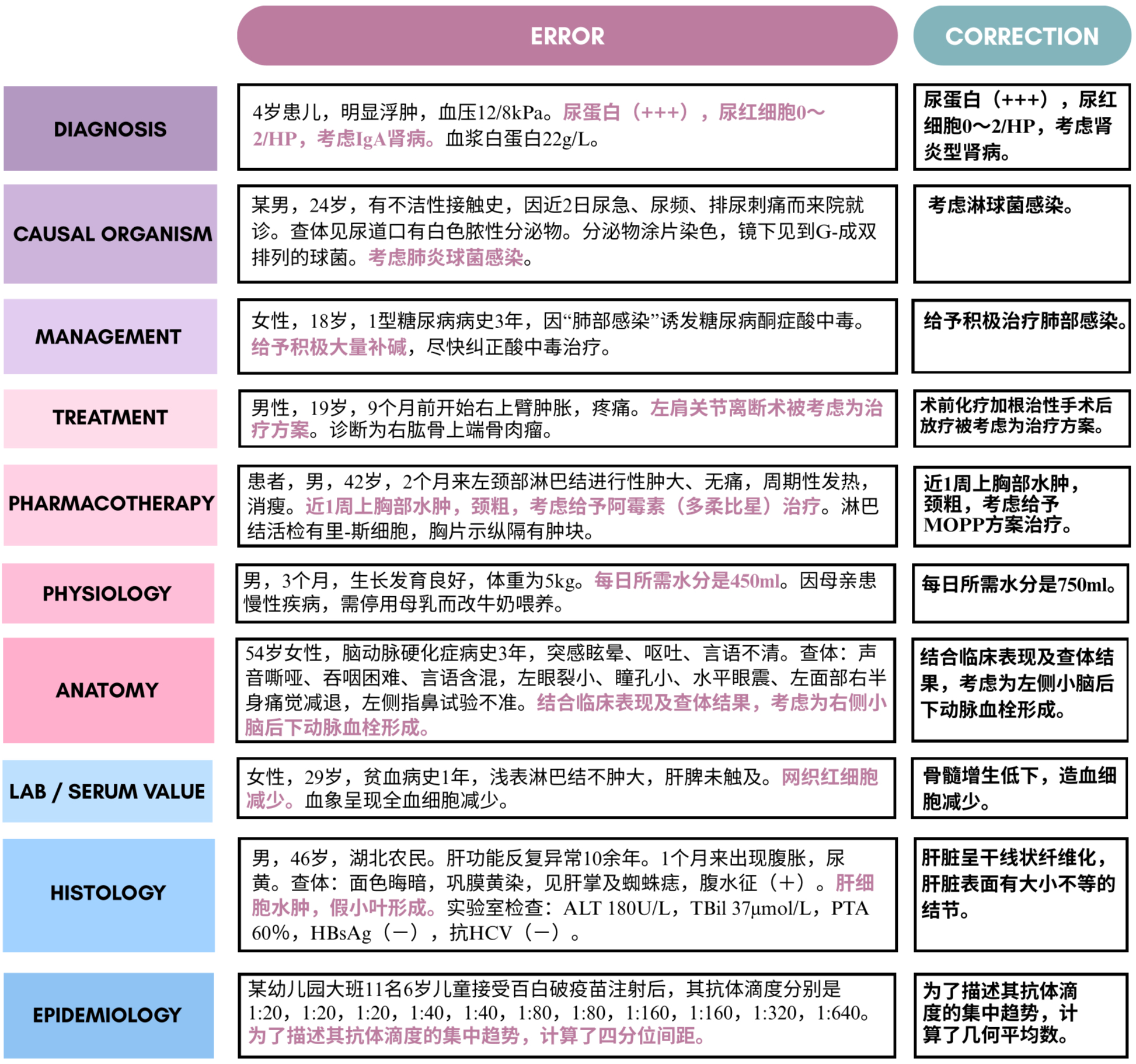}
    \vspace{-2mm}
    \caption{Illustrative Examples of the Ten Error Types Defined in the MedErrBench-CN.}
    \label{fig:Error_type_CN}
\end{figure*}

\begin{figure*}[t]
    \centering
    \includegraphics[width=1\textwidth,height=0.8\textheight,keepaspectratio]{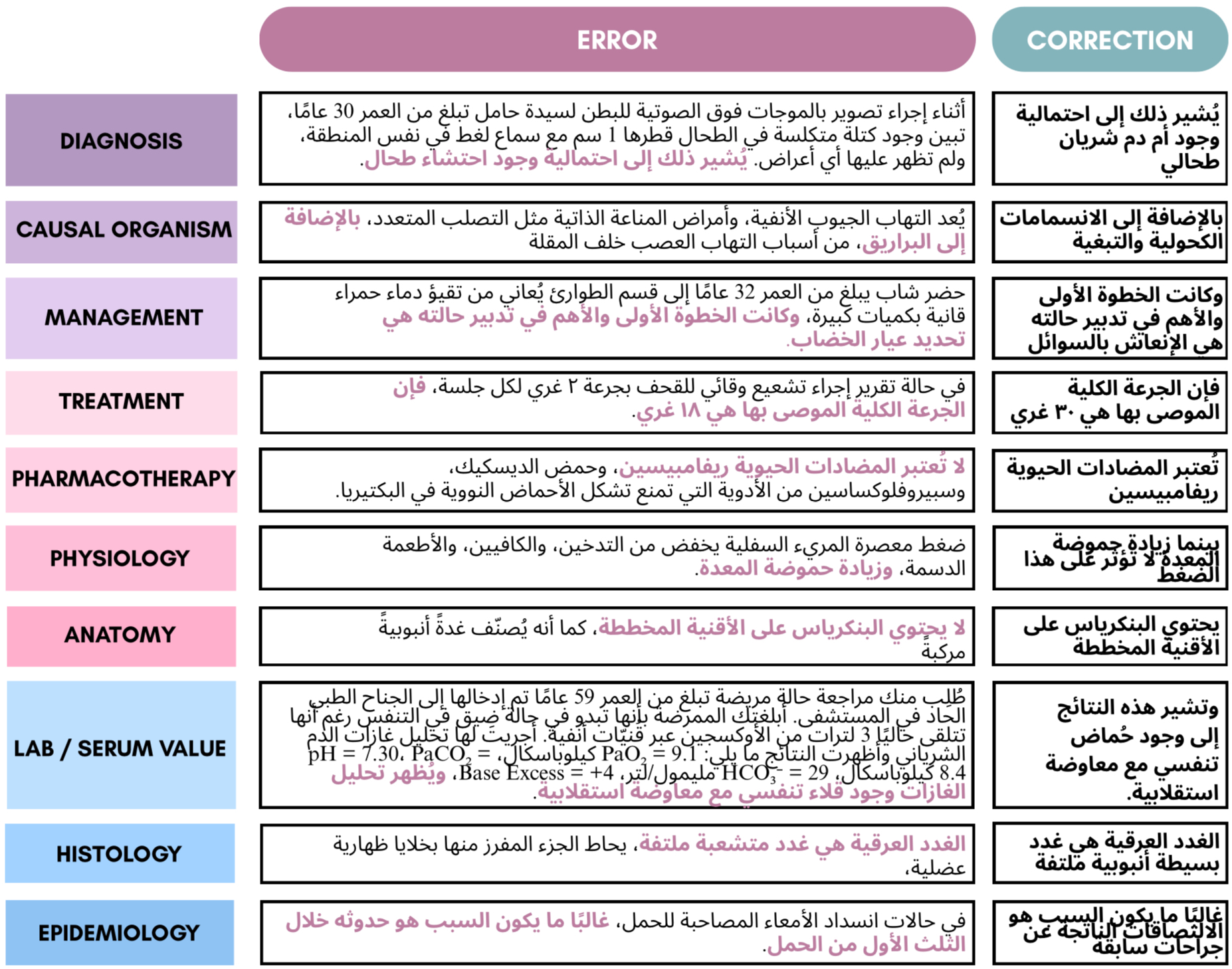}
    \vspace{-2mm}
    \caption{Illustrative Examples of the Ten Error Types Defined in the MedErrBench-Ara.}
    \label{fig:Error_type_Ara}
\end{figure*}

Figure \ref{fig:Error_type_En}, \ref{fig:Error_type_CN} and \ref{fig:Error_type_Ara} are the illustrative examples of the ten error types defined in the MedErrBench dataset.

\section{Performance of Existing Error Detection and Correction Methods on MedErrBench}

The MEDIQA-CORR 2024 shared task \cite{abacha2024overview} focused on detecting and correcting multiple types of medical errors in clinical texts and attracted participation from seventeen teams. We attempted to benchmark representative systems from this shared task on MedErrBench, our trilingual dataset. However, due to the unavailability or inactivity of most public GitHub repositories, only a limited subset of methods could be successfully reproduced and evaluated. Specifically, we benchmarked CLD-MEC \cite{alzghoul2024cld}, Maven \cite{jadhav2024maven}, and VerbaNexAI \cite{pajaro2024verbanexai}, including both VerbaNexAI-GRU and VerbaNexAI-ClinicalBERT.

The detailed results are reported in Table~\ref{tab:task-specific}. Overall, these task-specific error detection and correction models underperform compared to state-of-the-art language-specialized LLMs, suggesting that large pretrained models with strong multilingual and generative capabilities may generalize more effectively across diverse error types and languages than systems optimized for narrow task formulations.

Maven relies on a Retrieval-Augmented Generation (RAG) pipeline that is explicitly designed for English medical text. Its retrieval component uses an English-only knowledge base. When applied to Arabic or Chinese clinical texts, the retrieval mechanism fails to identify semantically relevant context, as embeddings derived from non-English inputs do not align well with English documents in the knowledge base. As a result, Maven produces irrelevant or incorrect disease or pathogen predictions and cannot be reliably evaluated on the Chinese or Arabic subsets of MedErrBench.

In addition, both VerbaNexAI-GRU and VerbaNexAI-ClinicalBERT do not produce localization or correction outputs, as they are fundamentally classification-based models rather than generative systems. Consequently, they are only partially comparable to generative LLM-based approaches in the full error detection and correction setting.

\renewcommand{\arraystretch}{0.85}
\begin{table*}[t]
\centering
\caption{Results on Error Detection and Correction Models.}
\label{tab:task-specific}
\resizebox{0.9\textwidth}{!}{%
\begin{tabular}{lcccccccc}
\toprule
\textbf{Models} & \multicolumn{1}{c}{\textbf{Detection}} & \multicolumn{1}{c}{\textbf{Localization}} & \multicolumn{6}{c}{\textbf{Error Correction}} \\
\textbf{} & \textbf{Accuracy} & \textbf{Accuracy} & \textbf{ROUGE-1} & \textbf{ROUGE-2} & \textbf{ROUGE-L} & \textbf{BLEU} & \textbf{BERTScore} & \textbf{BLEURT} \\
\midrule
\rowcolor{gray!15} \multicolumn{9}{c}{\textit{MedErrBench-EN}} \\
\midrule
CLD-MEC	&0.639	&0.615	&0.502	&0.477	&0.497	&0.462	&0.505	&0.534 \\
Maven	&0.264	&0.192	&0.035	&0.021	&0.027	&0.013	&0.128	&0.081 \\
VerbaNexAI-GRU	&0.510	&-	&-	&-	&-	&-	&-	&- \\
VerbaNexAI-ClinicalBERT 	&0.601	&-	&-	&-	&-	&-	&-	&- \\
\midrule
\rowcolor{gray!15} \multicolumn{9}{c}{\textit{MedErrBench-CN}} \\
\midrule
CLD-MEC	&0.655	&0.630	&0.479	&0.430	&0.472	&0.414	&0.581	&0.503 \\
\midrule
\rowcolor{gray!15} \multicolumn{9}{c}{\textit{MedErrBench-Ara}} \\
\midrule
CLD-MEC	&0.464	&0.381	&0.302	&0.276	&0.296	&0.260	&0.355	&0.315 \\
\bottomrule
\end{tabular}%
}
\end{table*}

\subsection{Cross-lingual Generalization}
\label{appendix: Cross-lingual}

\begin{figure*}[]
    \centering
\includegraphics[width=0.85\textwidth]{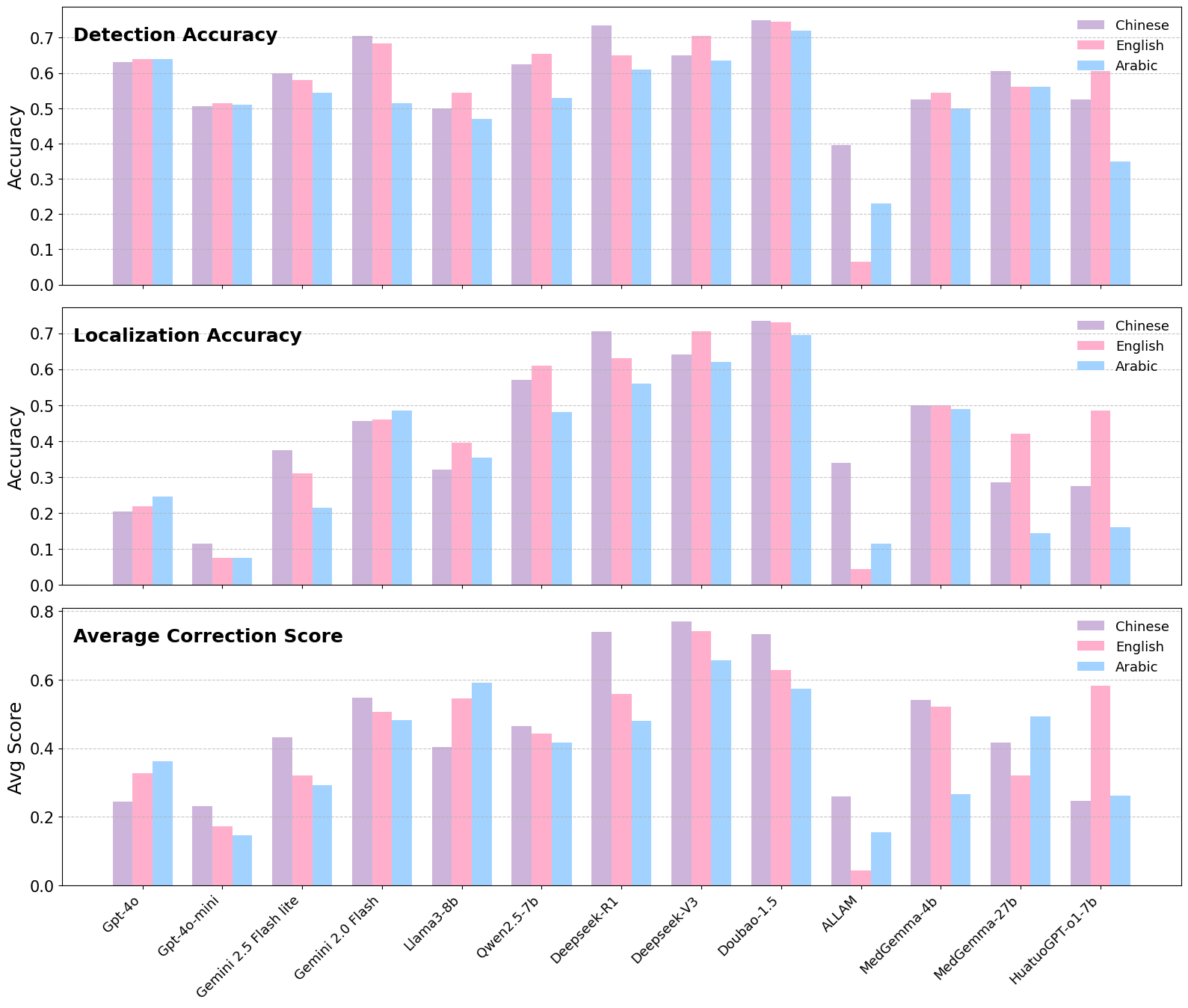}
    \vspace{-2mm}
    \caption{Cross-lingual performance comparison of LLMs on Chinese-origin tasks.} 
    \vspace{-4mm}
    \label{fig: Cross-lingual_Generalization}
\end{figure*}

To assess the effectiveness of machine-translated multilingual data for error detection, localization, and correction, we translated the Chinese dataset into English and Arabic. Chinese was selected arbitrarily as the source language and not for any specific preference, but to test whether linguistic representation differences across languages and cross-lingual variation alone affect model performance. The results are shown in Figure \ref {fig: Cross-lingual_Generalization}. We observe that across all three tasks, most models experience a performance drop to varying degrees on the translated English and Arabic datasets, with the decline being particularly pronounced on the Arabic data, especially with larger gaps observed in localization accuracy and average correction scores. This is mainly due to the substantial structural differences of the language and insufficient training data. In particular, translated data fails to capture the unique expressions and error patterns specific to Arabic, making it difficult for models to effectively transfer learning. Therefore, relying solely on translated data cannot meet the demands of high-quality multilingual models. It is crucial to construct authentic, diverse, and high-quality native multilingual datasets, so that the models deeply understand the characteristics of different languages, improve their capabilities in fine-grained tasks like localization and correction, and thereby enhance cross-lingual generalization performance.

\end{document}